%% file: main.tex
\documentclass[10pt]{article} 

\usepackage[preprint]{rlj} 

\usepackage{amssymb}            
\usepackage{mathtools}          
\usepackage{mathrsfs}           
\usepackage{graphicx}           
\usepackage{subcaption}         
\usepackage[space]{grffile}     
\usepackage{url}                
\usepackage{lipsum}             
\usepackage{booktabs}

\usepackage{cleveref}
\usepackage{ifthen}
\usepackage{csquotes}

\usepackage[noend]{algpseudocode}
\usepackage{algorithm}

\newcommand{\display}[3][normal]{%
  \begin{tabular}[c]{@{}c@{}}
    \ifthenelse{\equal{#1}{bold}}{\textbf{#2}}{%
      \ifthenelse{\equal{#1}{underline}}{\underline{#2}}{#2}%
    } \\[-3pt]
    \ifthenelse{\equal{#1}{bold}}{\footnotesize \textbf{$\pm$ #3}}{%
      \ifthenelse{\equal{#1}{underline}}{\footnotesize \underline{$\pm$ #3}}{\footnotesize $\pm$ #3}%
    }
  \end{tabular}%
}
\newcommand{\displayheader}[2]{%
  \begin{tabular}[c]{@{}c@{}}
    #1 \\[-3pt]
    #2
  \end{tabular}%
}

\crefname{table}{Table}{Tables}
\crefname{figure}{Figure}{Figures}
\crefname{equation}{Eq.}{Eqs.}
\Crefname{table}{Table}{Tables}
\Crefname{figure}{Figure}{Figures}
\Crefname{equation}{Eq.}{Eqs.}

\title{Multi-Agent Reinforcement Learning for Inverse Design in Photonic Integrated Circuits}

\setrunningtitle{MARL for Inverse Design in PICs}

\author{Yannik Mahlau\textsuperscript{1}, Maximilian Schier\textsuperscript{1}, Christoph Reinders\textsuperscript{1}, Frederik Schubert\textsuperscript{1}, Marco Bügling, Bodo Rosenhahn\textsuperscript{1}}

\emails{mahlau@tnt.uni-hannover.de}

\affiliations{
$^{1}$\textbf{Institute of Information Processing, Leibniz University Hannover, Germany}
}

\contribution{
We introduce the design of photonic integrated circuit components as a discrete optimization problem, which we implement as a multi-agent reinforcement learning (MARL) environment.
This bandit-like MARL environment tests the interaction of multiple thousand agents with very few samples.
}{
Photonic integrated circuits enable optical computing, which is a new field for fast and energy efficient hardware accelerators \citep{McMahon_2023}.
Previous research in MARL mostly focused on a handful of agents using millions of training samples to learn an environment with many states \citep{rutherford2023jaxmarl}.
In contrast, we introduce a bandit setting with a single environment state, but multiple thousands of agents using only 10000 training samples.
The sample efficiency is important because electromagnetic simulations for environment steps are time-consuming \citep{mahlau2024flexible}.
}

\contribution{
To solve the challenges of our new environment, we develop two multi-agent reinforcement learning algorithms.
They are based on proximal policy optimization \citep{schulman2017proximal} and an actor-critic approach with stochastic policies similar to the soft-actor-critic \citep{haarnoja2018soft}.
In extensive experiments, we show that our algorithms outperform previous state-of-the-art.
}{
Inverse design in photonics has previously almost exclusively been performed using gradient-based optimization \citep{schubertmahlau2025quantized}, which can quickly find a decent solution, but its susceptibility to getting stuck in local minima during optimization impedes performance.
}

\contribution{
We publish the reinforcement learning environment and training algorithms as open-source.
}{We hope to facilitate reproducibility and further research.}


\summary{
\input{sections/00_abstract}
}

\begin{document}

\makeCover  
\maketitle  

\begin{abstract}
\input{sections/00_abstract}\footnote{Our open-source implementation can be found at \url{https://github.com/ymahlau/blend}}
\end{abstract}

\input{sections/01_introduction}
\input{sections/02_background}

\input{sections/03_method}

\input{sections/04_experiments}

\input{sections/05_conclusion}

\subsubsection*{Broader Impact Statement}
\label{sec:broaderImpact}
The research presented in this paper advances inverse design methodologies for PIC components.
Although these innovations promise progress in optical computing and energy efficiency, we must carefully consider their potential impact on employment for human designers.
We believe that optimal design outcomes emerge from collaborative processes that combine human expertise with automated systems.
Critical to this approach is addressing questions of social acceptance and ensuring technological advancement proceeds by augmenting rather than replacing human capabilities.

\appendix





\subsubsection*{Acknowledgments}
\label{sec:ack}
This work was supported by the Federal Ministry of Education and Research (BMBF), Germany under the AI service center KISSKI (grant no. 01IS22093C), the Lower Saxony Ministry of Science and Culture (MWK) through the zukunft.niedersachsen program of the Volkswagen Foundation and the Deutsche Forschungsgemeinschaft (DFG) under Germany’s Excellence Strategy within the Cluster of Excellence PhoenixD (EXC 2122) and (RO2497/17-1). 
Additionally, this was funded by the Deutsche Forschungsgemeinschaft (DFG, German Research Foundation) – 517733257.


\bibliography{main}
\bibliographystyle{rlj}


\input{sections/06_supplementary}



\end{document}

%% file: sections/00_abstract.tex
Inverse design of photonic integrated circuits (PICs) has traditionally relied on gradient-based optimization.
However, this approach is prone to end up in local minima, which results in suboptimal design functionality.
As interest in PICs increases due to their potential for addressing modern hardware demands through optical computing, more adaptive optimization algorithms are needed.
We present a reinforcement learning (RL) environment as well as multi-agent RL algorithms for the design of PICs.
By discretizing the design space into a grid, we formulate the design task as an optimization problem with thousands of binary variables.
We consider multiple two- and three-dimensional design tasks that represent PIC components for an optical computing system.
By decomposing the design space into thousands of individual agents, our algorithms are able to optimize designs with only a few thousand environment samples.
They outperform previous state-of-the-art gradient-based optimization in both two- and three-dimensional design tasks.
Our work may also serve as a benchmark for further exploration of sample-efficient RL for inverse design in photonics.

%% file: sections/01_introduction.tex
\section{Introduction}
Modern computing and machine learning are fundamentally based on the representation and processing of information through digital electrical signals.
This approach has driven technological advancement for decades.
Although modern hardware has achieved significant improvements in computing power, particularly through parallel architectures like Graphics Processing Units (GPUs), fundamental limitations are being reached \citep{markov2014limits}.
Although the number of transistors continues to increase, the clock rate of individual processor cores has reached a plateau.
This constraint persists even as GPU architectures leverage massive parallelization to achieve higher computational throughput.
Consequently, this led to renewed consideration of analog computing for specialized applications \citep{haensch19, kazanskiy2022optical}.

In optical computing, the digital representation is replaced by an analog encoding using electromagnetic light waves.
This has the advantages of high bandwidth, operation speed, and energy efficiency \citep{McMahon_2023}.
Computation is performed on photonic integrated circuits (PIC), where optical components are connected for data input, output, and computation.
PICs are especially interesting for neural network inference, whose energy consumption has increased drastically in the last years \citep{DESISLAVOV2023100857}.
To illustrate the potential speedups, our designs perform a small scalar-vector multiplication in about 150 femtoseconds, which is about 2500 times faster than a single clock cycle of a classical electrical computer.

However, analog computing requires high accuracy to work well, as errors through multiple operations accumulate.
Designing a PIC component by hand is difficult, as designs are often counterintuitive and have a large number of parameters \citep{inverse_design_nano}.
Therefore, inverse design has to be used, where a design is optimized automatically using an electromagnetic simulation.
Since these simulations are differentiable, it has been very popular to optimize PIC components using gradient-based optimization \citep{inverse_design_nano}.
However, gradient-based optimizations often get stuck in local minima.
Therefore, it is necessary to find a better optimization algorithm.

\begin{figure}[!t]
    \centering
    \subfloat[Environment Scene]{
        \includegraphics[height=0.23\textwidth]{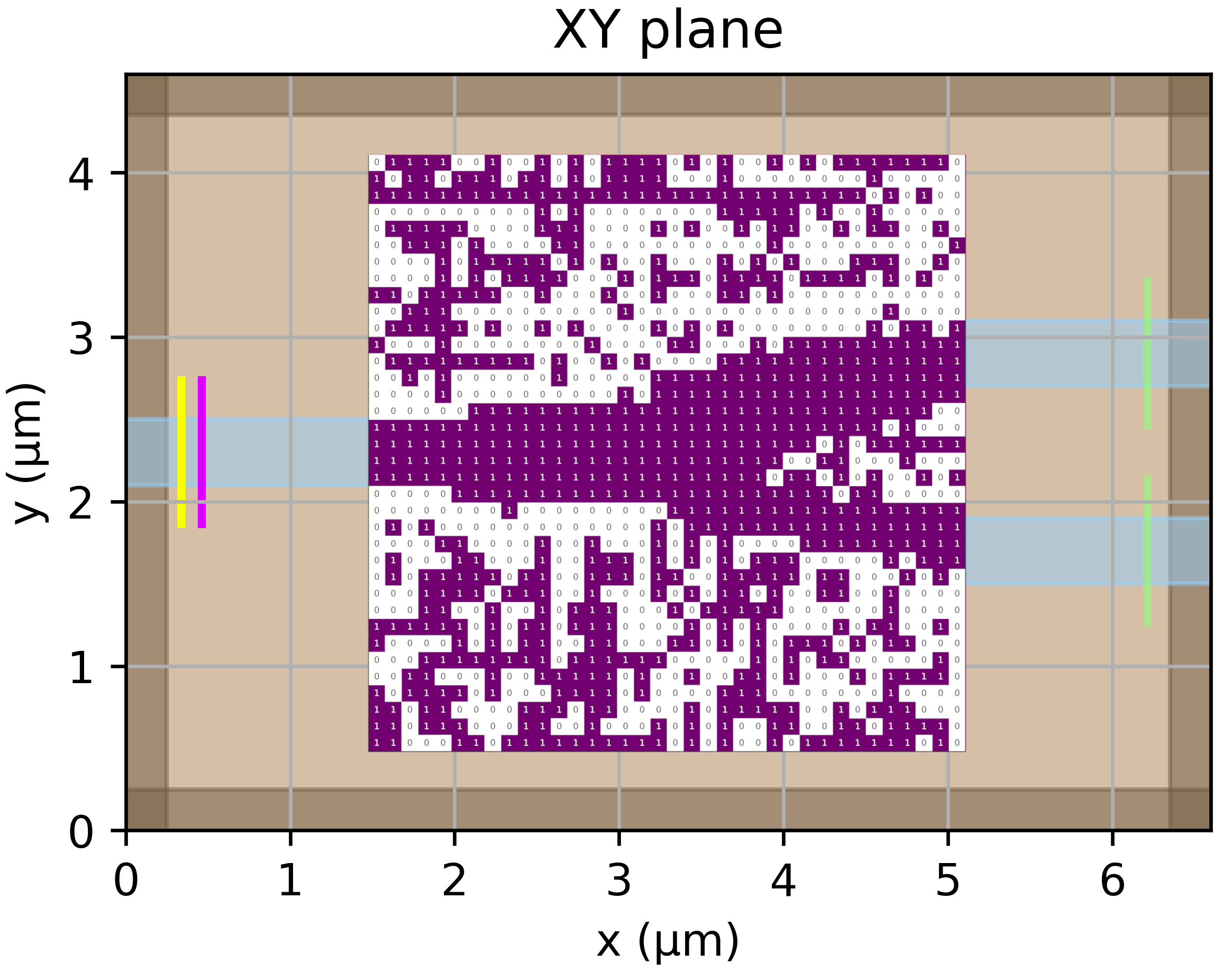}
    }
    \subfloat[Simulation]{
        \includegraphics[height=0.23\textwidth]{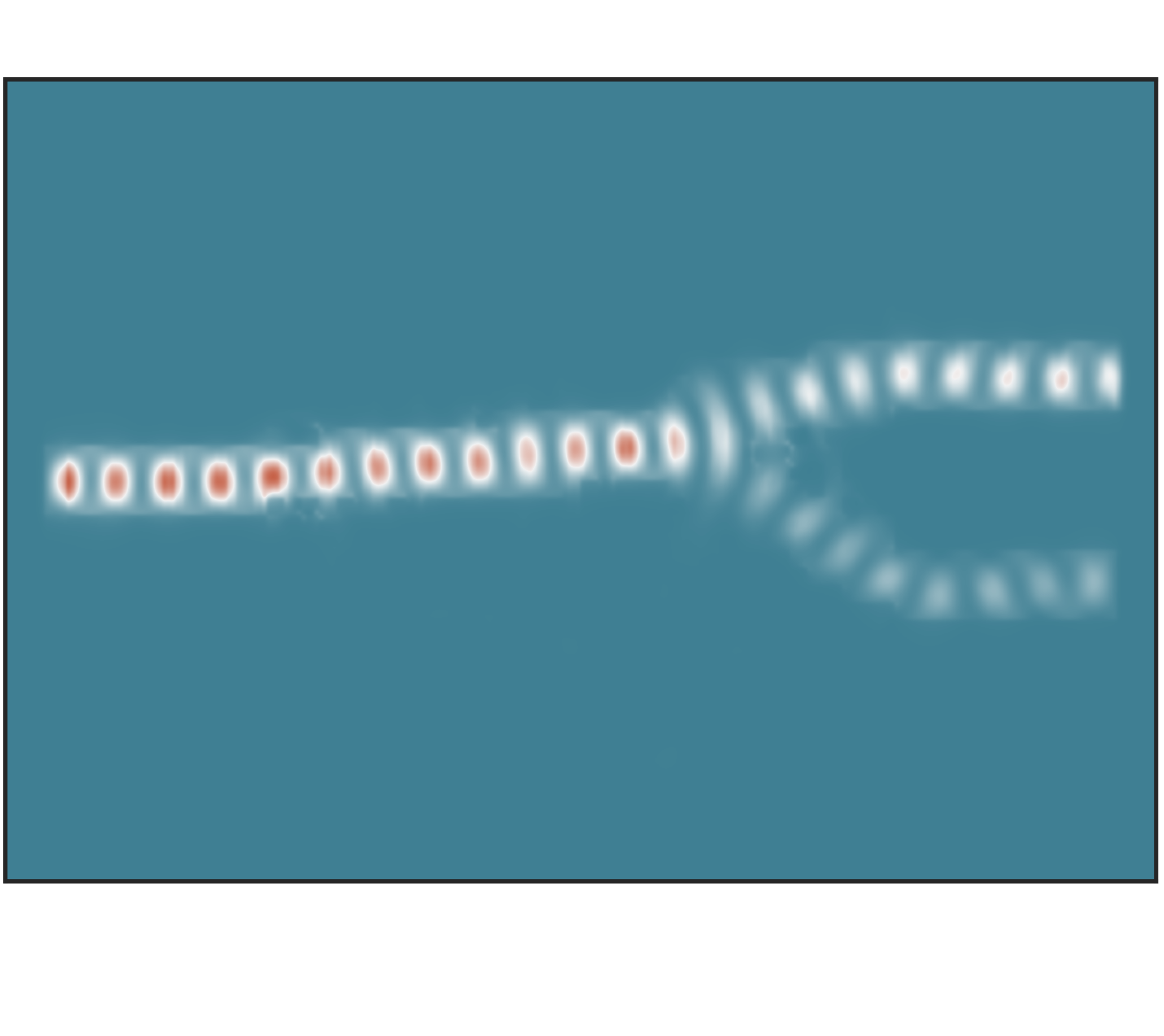}
    }
    \subfloat[Optimization Results]{
        \includegraphics[height=0.22\textwidth]{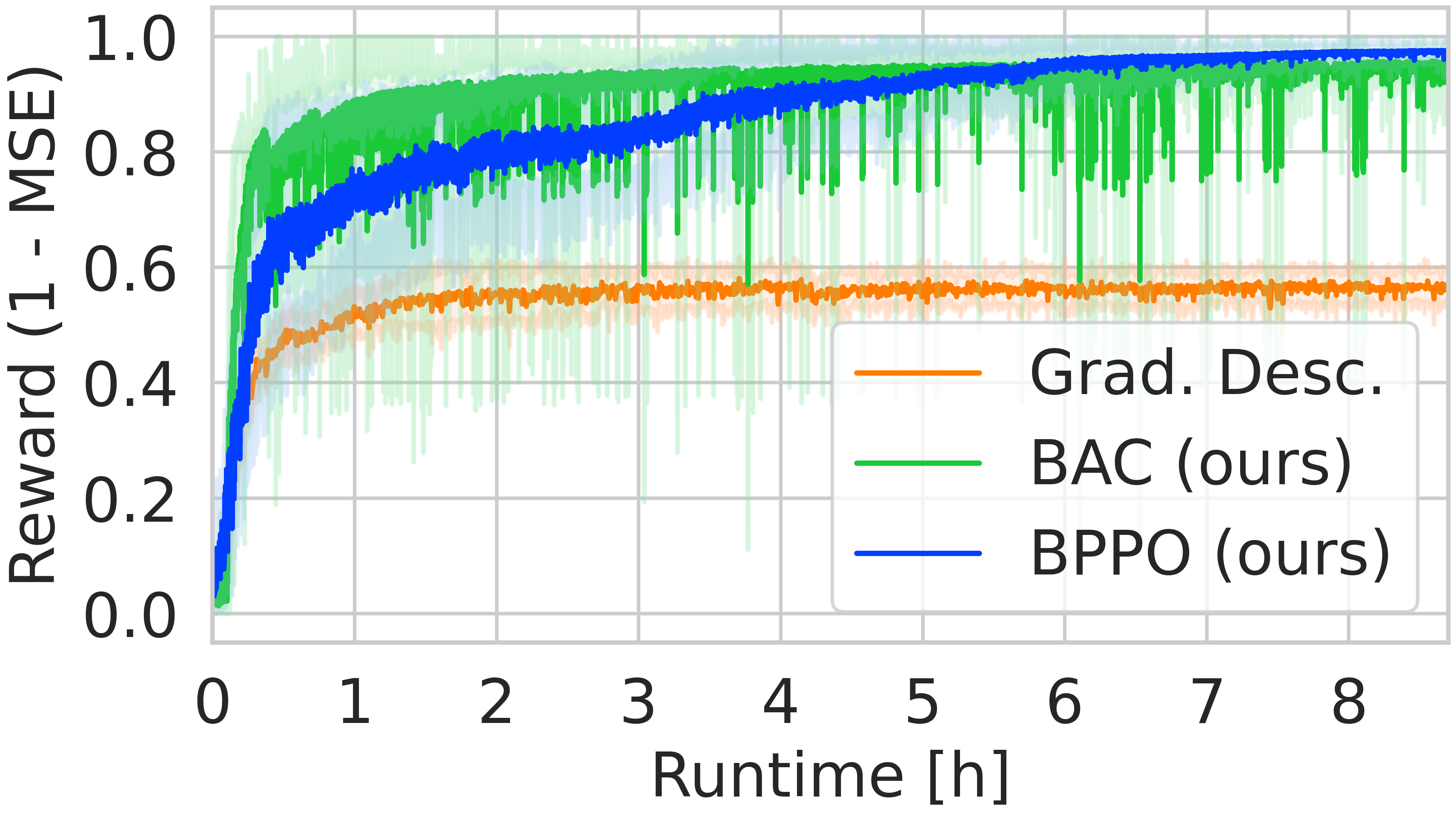}
    }
    
    \caption{
    Design task of a linear operation on a photonic integrated circuit.
    In (a), 65\% of the incoming light emitted by a source (yellow) in the left waveguide (blue) should be routed to the top right waveguide (blue), while 35\% of the light should go to the bottom right waveguide (blue).
    Transmission is measured as the ratio between output- (green) and input-detector (pink).
    The design task is a binary optimization problem for choosing silicon or air at every voxel.
    In (b), an electromagnetic simulation of the design is shown.
    During optimization (c), gradient descent gets stuck in a local minimum, while our BPPO and BAC show better exploration behavior.
    }
    \label{fig:teaser}
\end{figure}

We formulate the task of finding a good design for a PIC component as a discrete optimization problem.
By discretizing the design space, the task can be formulated as placing either material or air at every voxel in 3D space.
Since electromagnetic simulations are expensive, a learning algorithm must optimize a large design space using few samples.
In extensive experiments, we show that our new multi-agent RL algorithms are able to deal with these challenges through the decomposition of the action space into multiple agents.
The algorithms are based on proximal policy optimization \citep{schulman2017proximal} and an actor-critic approach with stochastic policies similar to soft actor-critic (SAC) \citep{haarnoja2018soft}.
Through multiple design tasks, we illustrate that both algorithms significantly outperform gradient-based optimization, which was previously state-of-the-art in inverse design.
In \cref{fig:teaser}, an example for optimizing a PIC design is shown.


%% file: sections/02_background.tex
\section{Background}
Inverse design of PIC components is based on electromagnetic simulations, which allows analysis by simulating light propagation through the component.
The Finite-Difference Time-Domain (FDTD) method \citep{taflove} is the most popular method for such a simulation \citep{dory2019inverse, augenstein20}.
Light is characterized by an electric field $E$ and magnetic field $H$, which are three-dimensional vectors at every point in space.
The propagation of light can be computed using Maxwell's equations \citep{maxwells_eq}
\begin{align}
    \frac{\partial H}{\partial t} = - \frac{1}{\mu} \nabla \times E \hspace{1cm} &\text{and} \hspace{1cm}
    \frac{\partial E}{\partial t} = \frac{1}{\varepsilon} \nabla \times H.
\end{align}
The electric and magnetic fields are updated in a leapfrog pattern, where the electric field is updated based on the magnetic field and vice versa.
To efficiently compute these updates, space and time are discretized according to the Yee grid \citep{kaneyeeNumericalSolutionInitial1966}.
Specifically, the electric field is defined on whole integer time steps on the edges between spatial grid points.
In contrast, the magnetic field is defined in between two integer time steps on the faces between four spatial grid points.
This complicated arrangement ensures that the curl operation ($\nabla \times$) can be computed quickly and accurately since no interpolation is necessary.

PICs are fabricated using either silicon or polymer materials.
These different fabrication methods admit different fabrication constraints.
Silicon PICs are manufactured in a subtractive process, resulting in 2D designs with uniform extrusion in the third axis \citep{HAN2014271, HUNG2002256}.
Although silicon can be fabricated at nanometer resolution \citep{cai19}, we restrict the resolution of our designs to an economically viable size of $80$nm.
In contrast to silicon, polymer can be fabricated into intricate 3D structures using the two-photon polymerization (2PP) process \citep{2pp_basics}.
But, 2PP has other design constraints.
Specifically, no material can float in the air, and a design cannot have enclosed air cavities.
Furthermore, the resolution of 2PP is more coarse than that of silicon fabrication with a minimum feature size of $500$nm.

\section{Related Work}

The application of reinforcement learning to photonic inverse design has been enabled by recent speedups in electromagnetic simulation \citep{mahlau2024flexible, tidy3d}.
Early work focused on 1D topologies, which inherently have a small design space.
\citet{jiang2021otf,jiang2022reinforcement} proposed combining unsupervised learning and RL with genetic algorithms to optimize multilayer solar absorbers.
Similarly, \citet{doi:10.1021/acsphotonics.1c00839} applied Deep Q-Networks to the design of 1D metasurfaces.
\citet{ParkKimJungParkSeoKimParkParkJang+2024+1483+1492} developed a combination of gradient-based optimization and Deep-Q learning to optimize 1D metagratings.
Furthermore, a great deal of work focused on optimizing a small number of parameters of a fixed shape parameterization \citep{LiZhangXieGongDingDaiChenYinZhang+2023+319+334, YU2025131402, Shams2024DeepTR, witt2023reinforcement}.
Some initial success in 2D designs was achieved by \citet{Butz:23}, which optimized a mode converter with 2070 binary variables using an undisclosed RL algorithm.

To the best of our knowledge, large-scale 2D or fully 3D topology optimization has only been achieved using gradient-based optimization with the adjoint method \citep{dory2019inverse, Mansouree_20}, automatic differentiation \citep{schubertmahlau2025quantized, hughes2019forward, Tang2023TimeRD}, or a combination of both \citep{luce2024merging}.

%% file: sections/03_method.tex
\section{Reinforcement Learning for Inverse Design}
We model the problem of designing a PIC component as a discrete optimization problem.
For every voxel in the discretized design space, the designer has to make a decision wether to place material or leave this spot empty, i.e. place air.
The discretization of the design space in a grid follows naturally from the discretization of the FDTD simulation.
Mathematically, the design space is $\mathcal{A} = \{0, 1\}^N$, where $N$ is the number of discretized voxels.
We denote a joint action using the bold letter $\mathbf{a} = (a_1, \ldots, a_N)$.
The objective is to find the best joint action $\mathbf{a}^* = \operatorname{arg}\operatorname*{max}_{\mathbf{a} \in \mathcal{A}} R(\mathbf{a})$, where $R : \mathcal{A} \rightarrow \mathbb{R}$ is the payoff function of the environment.
The payoff function performs an FDTD simulation, which is expensive.
Therefore, we define a budget $T$ that specifies how often the payoff function can be queried during optimization.
This formulation can be viewed as a multi-armed bandit, except in contrast to classical formulations, performance is only measured using the best reward, i.e. $\max_{t \in \{1, \ldots, T\}}R(\mathbf{a}_t)$.


\subsection{Baseline Optimization Algorithms}
In the past, most inverse design has been performed using gradient-based optimization, which calculates the gradient through the differentiable FDTD simulation and performs supervised optimization.
During optimization, the material permittivity $\varepsilon$ is modeled as a continuous parameter.
For simulation, the continuous permittivity is mapped to the closest material permittivity.
This mapping ensures that the simulation is physically valid at every optimization step.
Furthermore, it enforces fabrication constraints for polymer designs by removing floating material and filling enclosed cavities.
However, the mapping introduces a non-differentiable operation, but this issue can be overcome with a straight-through estimator \citep{schubertmahlau2025quantized}.

Nevertheless, gradient computation is costly in electromagnetic simulations and optimization is prone to get stuck in local minima.
An optimization procedure that does not require gradients is the evolutionary algorithm \citep{Jin2003}.
In this algorithm, a population of random binary designs is initialized, which are randomly mutated and recombined based on their performance during optimization.
Another well-studied optimization procedure from the bandit literature is the upper confidence bound.
Specifically, the decoupled upper confidence bound for trees (DUCT) is often used to decompose the large joint action space of a multi-agent system into small individual action spaces \citep{tak_sm_mcts, MahSch2024a}.
This idea can be applied here, so that every voxel individually keeps track of the rewards associated with placing material or air.
Every voxel then individually chooses to place material or air in the next iteration based on the DUCT formula
\begin{align}
a_n^* = {\arg\max}_{a \in \{0, 1\}} \frac{w_n^a}{v_n^a} + c \cdot \frac{\sqrt{v_n^0 + v_n^1}}{v_n^a},
\end{align}
where $v^a$ is the number of times action $a$ has been used and $w^a$ is the accumulated sum of rewards.
The exploration constant $c$ is a hyperparameter that balances between exploration and exploitation.

\subsection{Multi-Agent Reinforcement Learning for Bandits}
We adapt two different reinforcement learning algorithms to the bandit setting of inverse design.
Both algorithms make use of the same neural network architecture, where the parameters are shared between all agents.
The input of the neural network is a positional encoding $O : \{1,\ldots,N\} \rightarrow \mathcal{O}$, mapping the agent index to an encoding providing a structural bias \citep{vaswani2017attention}.
In addition, agents implicitly share information through the shared neural network architecture.

\subsubsection{Bandit Actor-Critic (BAC)}

We implement a novel actor-critic approach with stochastic policies similar to SAC \citep{haarnoja2018soft} for the multi-agent bandit problem.
Let $\pi : \mathcal{O} \rightarrow \mathcal{P}(\{0, 1\})$ be a stochastic policy conditioned on a positional encoding.
We denote a single sampled action as $a_n \sim \pi(\cdot \,|\, O(n))$ and use $\mathbf{a} \sim \pi$ as the shorthand for a sampled joint action.
Thus, the optimization objective becomes $\pi^* = \operatorname{arg}\operatorname*{max}_{\pi} \operatorname*{\mathbb{E}}_{\mathbf{a} \sim \pi}[R(\mathbf{a})]$.
Next, we introduce a centralized critic $\mathbf{C}_{\psi} : \mathcal{A} \rightarrow \mathbb{R}$, which predicts the expected payoff of a joint action.
Its parameters $\psi$ are learned through gradient descent on the regression error of $\mathbf{C}$ with
\begin{align}
    J_{\mathbf{C}}(\psi) = \operatorname*{\mathbb{E}}_{(\mathbf{a}, r) \sim \mathcal{D}}\left[(r - \mathbf{C}_{\psi}(\mathbf{a}))^2\right]\, \text{.} \label{eq:bac_critic}
\end{align}
Since $\mathbf{C}$ is a differentiable surrogate of $R$, the parameters $\theta$ of a parameterized policy $\pi_{\theta}$ can be learned by gradient ascent on $\mathbf{C}$ using the objective
\begin{align}
    \label{eq:bac_pi}
    J_{\pi}(\theta) = \operatorname*{\mathbb{E}}_{\mathbf{a} \sim \pi_{\theta}}[\mathbf{C}_{\psi}(\mathbf{a})]\, \text{.} 
\end{align}
However, the sampling of $\mathbf{a} \sim \pi_{\theta}$ is not differentiable, since the action space is (multi-) discrete.
For single agent RL, this is commonly solved by expressing the expectation as a sum $\mathbb{E}_{\mathbf{a} \sim \pi}[\mathbf{C}(\mathbf{a})] = \sum_{\mathbf{a} \in \mathcal{A}}\pi(\mathbf{a}) \mathbf{C}(\mathbf{a})$ \citep{christodoulou2019soft,vieillard2020munchausen}.
But in our bandit setting, this is not tractable due to the large joint action space of size $\lvert \mathcal{A} \rvert = 2^N$.
We also considered calculating the closed form for any single agent while keeping the actions of other agents fixed, as done in MARL algorithms like COMA \citep{foerster2018counterfactual}.
But for such an objective, the number of critic evaluations would scale linearly with the number of agents, which becomes excessive for tens of thousands of agents.
Instead, we approximate the gradient of the expected value by straight-through estimation on one drawn sample \citep{bengio2013estimating}.
Specifically, for any action $a_n \sim \pi_{\theta}(\cdot \,|\, O(n))$, we approximate the gradient as $\nabla a_n \approx \nabla \pi_{\theta}(a_n=1 \,|\, O(n))$.
Thus, we can calculate the gradient through all agents using a single evaluation of the critic.

We present a short overview of BAC in Algorithm \ref{alg:bac}.
In contrast to other actor-critic frameworks, we use a large number of gradient steps on the policy and critic networks per collected sample from the environment.
For general Markov decision processes this would lead to estimation biases due to the temporal difference learning \citep{chen2021randomized}, but in the bandit setting this is not a concern.
However, in the bandit setting, the lack of stochasticity may be problematic.
The only stochasticity induced into the policy during optimization with Eq.~\eqref{eq:bac_pi} is through action sampling.
The loss of entropy of the policy through optimization leads to a loss of entropy of the gradient ascent, which can impede training performance \citep{amir2021sgd}.
We solve this in two ways.
Firstly, we regularly reinitialize the policy to carry out a completely new gradient ascent starting from a policy with high entropy.
Secondly, we mask a fraction $m$ of all agents for the gradient calculation in every gradient step.
Thus, the optimization would retain randomness even with a fully deterministic policy.

\algdef{SxnE}[REPEATN]{RepeatN}{EndRepeatN}[1]{\algorithmicrepeat\ #1 \textbf{times}}
\begin{algorithm}[tb]
\caption{Bandit Actor-Critic}\label{alg:cap}
\label{alg:bac}
\begin{algorithmic}[1]
\Require Simulation budget $T$, Critic gradient steps $U$, Policy gradient steps $G$, Critic learning rate $\lambda_C$, Policy learning rate $\lambda_{\pi}$, Critic batch size $B$
\State $\mathcal{D} \gets \{\}$
\State \text{Randomly initialize} $\psi, \theta$
\RepeatN{$T$}
    \State $\mathbf{a} = (a_1, \ldots, a_N)$, \text{with} $a_i \sim \pi_{\theta}(\cdot | O(i))$ \Comment{Sample from current policy}
\State $r \gets R(\mathbf{a})$ \Comment{Run simulation and observe pay-off}
\State $\mathcal{D} \gets \mathcal{D} \cup \{(\mathbf{a}, r)\}$ \Comment{Record experience}
    \RepeatN{$U$} \Comment{Train critic}
        \State $\psi \gets \psi - \lambda_C \nabla J_{\mathbf{C}}(\psi)$ \text{using a random batch of size $B$ from $\mathcal{D}$} \Comment{See Eq.~\eqref{eq:bac_critic}}
    \EndRepeatN
    \State \text{Randomly initialize} $\theta$
    \RepeatN{$G$} \Comment{Find new best policy}
        \State $\theta \gets \theta + \lambda_{\pi} \nabla J_{\pi}(\theta)$ \Comment{See Eq.~\eqref{eq:bac_pi}}
    \EndRepeatN
\EndRepeatN
\end{algorithmic}
\end{algorithm}

\subsubsection{Bandit Proximal Policy Optimization (BPPO)}
Proximal Policy Optimization (PPO) \citep{schulman2017proximal} has been one of the most successful reinforcement learning algorithms in recent years.
However, basic PPO would be difficult to apply here because of the large action space and small number of samples.
Following the idea of IPPO \citep{ippo} and MAPPO \citep{mappo} to decompose the large action space into multiple agents, we implement Bandit Proximal Policy Optimization (BPPO).
In the bandit setting, there is no notion of states, such that the PPO loss function for a single action $a_n$ can be written as 
\begin{align*}
    J(a_n, \theta_\text{old}, \theta) = \min \Bigl( \rho(a_n, \theta_\text{old}, \theta)  A^{\pi_{\theta_{\text{old}}}}(a_n), \, \text{clip} \bigl( \rho(a_n, \theta_\text{old}, \theta), 1 - \epsilon, 1 + \epsilon \bigr) A^{\pi_{\theta_{\text{old}}}}(a_n) \Bigr) ,
\end{align*}
where $\rho\left(a_n, \theta_\text{old}, \theta\right) = \frac{\pi_\theta\left(a_n | \,O\left(n\right)\right)}{\pi_{\theta_{\text{old}}}\left(a_n |\, O(n)\right)}$ is the policy ratio and $A^{\pi_{\theta_{\text{old}}}}(a_n)$ an advantage estimate for action $a_n$.
In contrast to classical PPO, the advantage estimate is not dependent on any state, such that we can remove the critic completely and estimate the advantage as
\begin{align*}
A^{\pi_{\theta_{\text{old}}}}(a_n) = r_{a_n} - \operatorname*{\mathbb{E}}_{\mathbf{a} \sim \pi_{\theta_{\text{old}}}}\bigl[R(\textbf{a})\bigr],
\end{align*}
where $r_{a_n}$ is the payoff from a single sample associated with playing action $a_n$.
The expected payoff of the old policy can be estimated using samples collected during rollout.

\begin{figure}[tb]
    \centering
    \includegraphics[width=\textwidth]{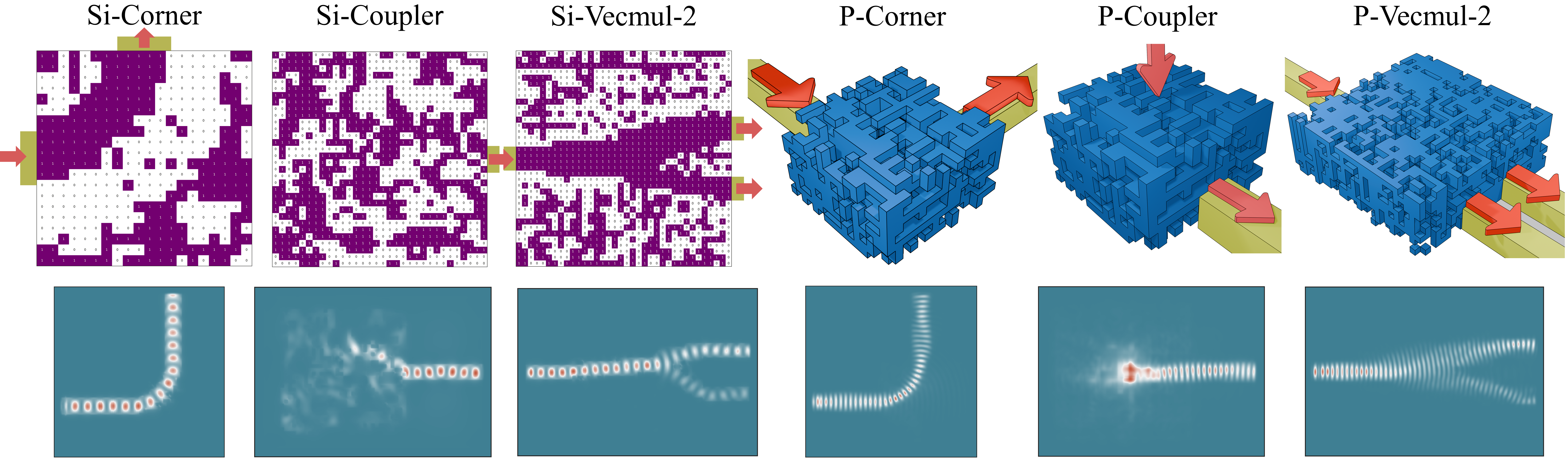}
    \caption{
    Optimized designs for the three different environments using either 2D silicon (purple) or 3D polymer designs (blue). The input and output waveguides are marked in green and with arrows. In the coupler environment, the input light comes from the top. The bottom row shows electromagnetic simulations of the designs.
    }
    \label{fig:env_designs}
\end{figure}

\subsection{Environment Design}

We introduce an reinforcement learning compatible environment for the design of PIC components.
It includes three different scenarios, covering all of the major components necessary to build an optical computing system.
The three different scenarios can be realized using silicon or polymer.

In the first scenario, light needs to be transferred from an optical fiber to the chip as data input.
Therefore, the challenge is to design a coupling element, which transfers light from free space into a waveguide.
The fiber is placed vertically above the respective coupling element.
The objective of the design is to transfer as much light as possible into the waveguide.
This transfer can be measured using the poynting flux $P = E \times H$, which intuitively represents how much energy flows in a specific direction.
Specifically, we would like to maximize the fraction of $P_x$ in the waveguide to $P_z$ below the source.
Due to the time reversibility of Maxwell's equations, this design could also be used to transfer light from a waveguide into free space to measure computational output.

Secondly, light needs to be routed on the optical chip.
Waveguides have low transmission loss on an optical chip as long as the waveguide is straight.
However, for complex optical chips, it may be necessary to use sharp bends in the waveguides for efficient routing.
These sharp bends introduce high transmission losses if the design is a simple round curve \citep{SnyderLove_1983}.
Therefore, the second task is to design a component that connects two waveguides at a $90^{\circ}$ angle.
Again, the objective is to transfer as much light as possible measured as the fraction of poynting flux.

Lastly, a basic computational component is necessary to actually perform a calculation.
Since we consider the standard linear form of Maxwell's equations, the computations are also restricted to linear operations.
This restriction could be alleviated using nonlinear materials, which we leave to future work because it is a topic of active research in the photonics community \citep{BogdanovMakarovKivshar2024}.
For simplicity, we assume that the data is represented as directional energy in the waveguide, i.e. poynting flux.
The simplest building block for a linear operation is a scalar-vector multiplication with a fixed vector, for example the trained weights of a neural network.
By chaining multiple scalar-vector multiplications, one could also perform more complicated linear operations like a matrix-vector or even matrix-matrix multiplication.
The scalar-vector multiplication can be implemented by distributing light from an input waveguide to multiple output waveguides according to the fixed vector.
We measure the quality of a design as $1 - \text{MSE}$, where MSE is the mean squared error between the desired and actual poynting flux at the output waveguides.
In \cref{fig:env_designs}, designs of the six different setups are shown and \cref{fig:teaser} shows more detailed analysis of the scalar-vector multiplication environment.

%% file: sections/04_experiments.tex
\begin{table}[!t]
\centering
\begin{tabular}{l|c|ccccccc}
\toprule
Environment & \#Agents & Random  & DUCT & Grad  & EA & IQL & \displayheader{BAC}{(ours)} & \displayheader{BPPO}{(ours)}\\
\midrule
Si-Corner & 400 & \display{27.5}{3.6} & \display{47.1}{12.1} & \display{74.4}{2.5} & \display{77.6}{6.2} & \display{80.9}{2.3} & \display[underline]{88.6}{0.8} & \display[bold]{91.7}{0.6}\\
Si-Coupler & 1024 & \display{9.7}{3.2} & \display{5.4}{0.5} & \display[underline]{41.5}{2.1} & \display{27.0}{2.3} & \display{13.0}{4.6} & \display{17.8}{0.8} & \display[bold]{51.6}{5.4}\\
Si-VecMul-2 & 1296 & \display{24.8}{2.3} & \display{44.2}{6.4} & \display{61.2}{3.0} & \display{73.60}{3.00} & \display{74.2}{4.4} &  \display[underline]{96.1}{0.4} & \display[bold]{97.7}{1.4}\\
Si-VecMul-5 & 4356 & \display{7.4}{0.7} & \display{3.9}{0.4} & \display{34.7}{1.2} & \display{35.2}{2.9} &  \display{63.1}{1.8} & \display[bold]{86.2}{4.1} & \display[underline]{76.2}{13.6}\\
\midrule
P-Corner & 2560 & \display{3.9}{0.95} & \display{3.7}{0.4} & \display{7.1}{2.5} & \display{31.7}{2.9} & \display{13.1}{14.9} & \display[underline]{51.3}{2.4} & \display[bold]{55.9}{2.6}\\
P-Coupler & 6912 & \display{1.7}{0.13} & \display{1.8}{0.08} & \display{8.0}{1.7} & \display{21.5}{3.5} & \display{5.5}{4.7} & \display[bold]{37.0}{3.1} & \display[underline]{33.2}{8.3}\\
P-VecMul-2 & 7840 & \display{2.7}{0.49} & \display{6.4}{0.3} & \display{12.3}{1.5} & \display{43.6}{4.7} & \display{77.3}{5.0} & \display[bold]{95.6}{0.8} & \display[underline]{92.2}{7.2}\\
P-VecMul-5 & 27040 & \display{0.3}{0.0} & \display{0.7}{0.02} & \display{6.1}{0.4} & \display{5.3}{1.1} & \display{53.7}{1.4} & \display[bold]{89.0}{3.3} & \display[underline]{69.8}{24.1}\\
\bottomrule
\end{tabular}
\caption{
Performance comparison across different environments and algorithms.
For the corner and coupler environments, performance is the transmission efficiency, while for the scalar-vector multiplication performance is measured as $1 - \text{MSE}$.
Mean and standard deviation are calculated over 5 seeds.
The best performing algorithm is highlighted as \textbf{bold} and the second best is \underline{underlined}.
}
\label{tab:performance_comparison}
\end{table}

\section{Experiments}

We test the algorithms introduced above for the different PIC-components.
As optimization in these environments is quite costly due to electromagnetic simulations, we devised a simple environment to optimize the various hyperparameters of all the algorithms presented above.
The objective of this testing environment is to find a stable initial condition for Conways game of life \citep{gardner1970mathematical}, which is a 2D binary grid optimization similar to our environments.
As this environment is quick to evaluate, we optimized all hyperparameters using this environment with Optuna \citep{akiba2019optuna}.
Only the gradient-based optimization cannot be optimized with this environment, as it is non-differentiable.
Gradient-based optimization has the learning rate as its only hyperparameter, which we optimized by a sweep over the silicon coupler environment.

In addition to the algorithms presented above, we also tested independent Q-learning (IQL), which applies the standard Q-learning approach to each agent individually \citep{tan1993multi, rutherford2023jaxmarl}.
In \cref{tab:performance_comparison}, the results of the evaluation are displayed.
DUCT performed only little better than random search in most environments, as it lacks coordination between different agents.
Gradient-based optimization performed better for the 2D designs of silicon than for the 3D polymer designs.
In 3D designs, the mapping of latent parameters to a physically valid design can introduce gradient errors by the straight-through estimator.
For example, when large parts of the design float in the air and are removed by the mapping, the gradient is calculated for a design with a large distance from the latent parameters.
Evolutionary algorithms (EA) perform similar to gradient-based optimization with better results in some environments and worse results in other environments.
The large number of agents and the small number of samples prevent EA from discovering an optimal solution.
IQL performs better than gradient-based optimization in most environments, but the epsilon-greedy exploration sometimes leads to suboptimal exploration behavior.
Our BAC and BPPO algorithms perform best, often with little difference.
However, BAC seems to perform better in environments with many agents, while BPPO performs better in smaller environments.
This indicates that the off-policy approach scales better to large designs, since all previous experience is contained in the replay buffer.
The designs optimized by BPPO are shown in \cref{fig:env_designs}.

For the corner environments, we can also compare the results against a naive circular corner design \cite{Bahadori_19}.
Measurement of this naive design in our environment yields a transmission of 88.4\%  for the silicon corner and 18.1\% for polymer.
In both cases, gradient-based optimization is unable to beat this simple baseline, while BAC and BPPO achieve better results.

\begin{figure}[!t]
    \centering
    \subfloat[Design Variance]{
        \includegraphics[width=0.48\textwidth]{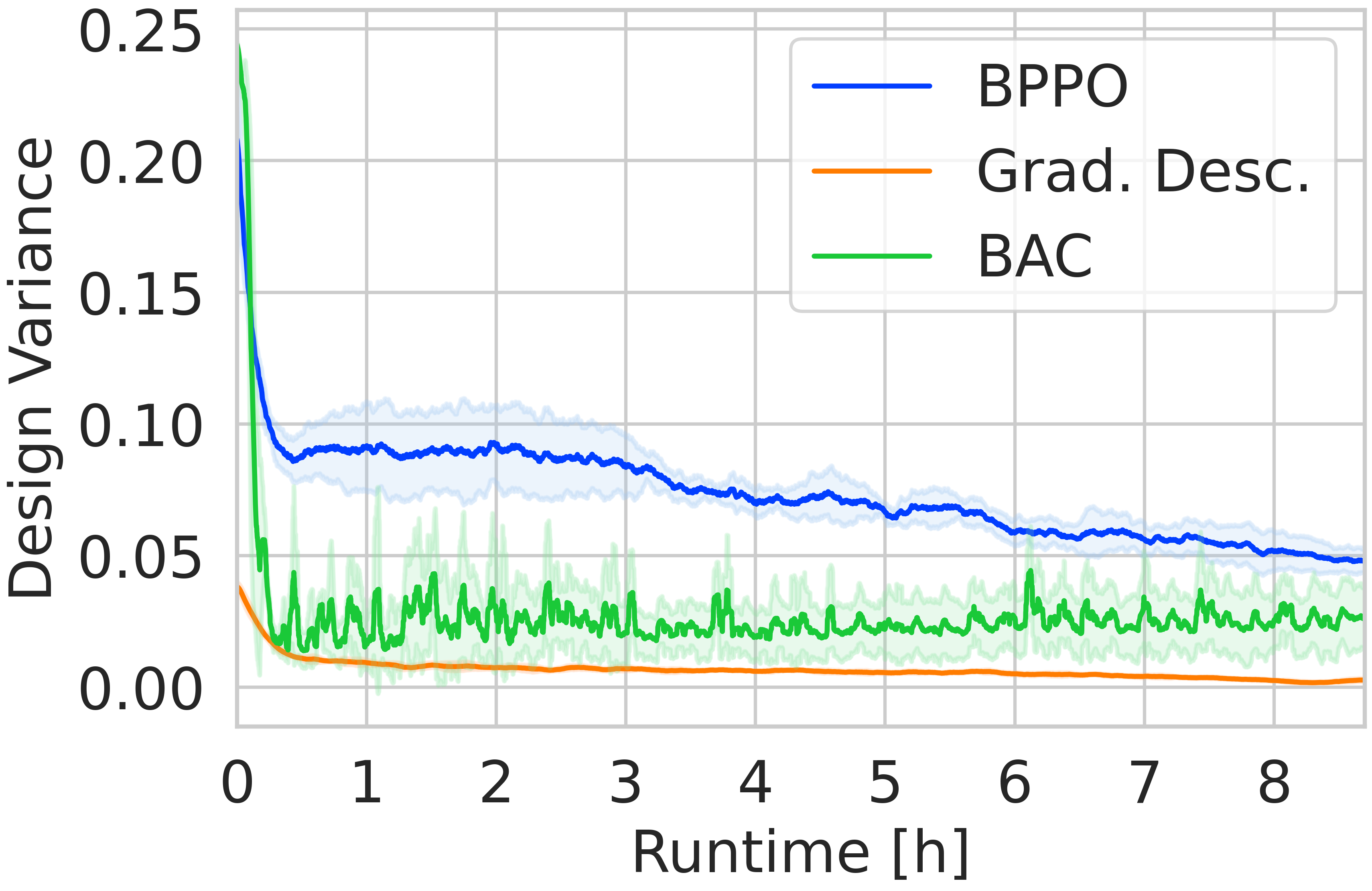}
    }
    \hfill
    \subfloat[Design Robustness]{
        \includegraphics[width=0.48\textwidth]{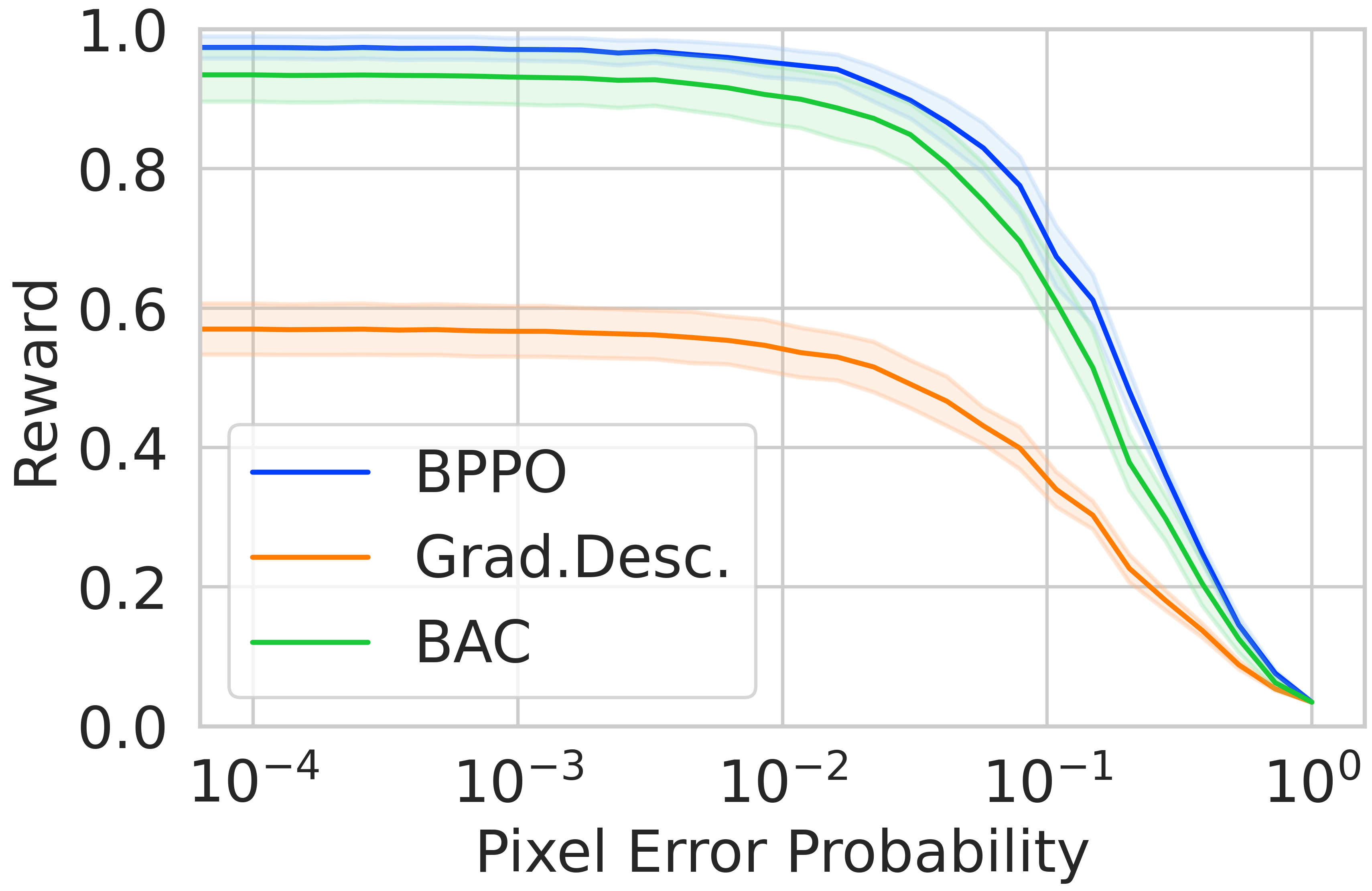}
    }
    \caption{
    Comparison of BPPO, BAC and gradient descent regarding design variance and robustness in the Si-Vecmul2 environment.
    In (a), the variance is calculated over a window of 50 time steps during optimization.
    In (b), voxels are set to a random binary value with varying error probability.
    }
    \label{fig:analysis}
\end{figure}

To show that gradient descent gets stuck in local optima, we analyze the variance of designs during optimization in \cref{fig:analysis}.
During optimization, the variance of designs produced by gradient descent quickly decreases, indicating that the optimization gets stuck a local optimum.
In contrast, BPPO and BAC have higher design variance than gradient descent, indicating better exploration behavior.
Another important consideration is the robustness of the designs produced.
For example, a high-performing design, whose performance collapses with a single voxel error, would be of little use in practice due to fabrication imperfections.
For BPPO and BAC this is not the case as their stochastic policies implicitly optimize for randomness in the design voxels.
Even if $10\%$ of the design voxels are randomly resampled, both algorithms still outperform the error-free designs of gradient-descent.

%% file: sections/05_conclusion.tex
\section{Conclusion and Future Work}

We developed a formulation of the inverse design task for PIC components as a discrete optimization problem.
For this bandit-like problem, we implemented environments representing the basic three components necessary to build an optical computing system.
These components can be fabricated with either silicon or polymer, which leads to 2D or 3D design tasks respectively.
The previous state-of-the-art gradient-based optimization can solve 2D silicon design tasks fairly well.
However, we showed that it does not produce optimal results because it is prone to get stuck in local optima.
Additionally, gradient descent struggles with 3D designs for polymer PICs.
In contrast, our new BAC and BPPO algorithms show better exploration behavior, resulting in better performing designs.

In future work, we plan to extend the framework to nonlinear materials, which would alleviate the restrictions of Maxwell's linear equations.
This would greatly increase the number of applications, for example building hardware accelerators for a trained neural network.
Furthermore, there exist technologies for multi-material fabrication of polymer \citep{HU2022102575}.
Extending the action space from a binary choice to a class of three or more materials would be another extension of our framework that needs to be analyzed.
Moreover, although our new algorithms outperform classical optimization algorithms by a large margin, they do not achieve perfect scores on our benchmarks.
For building a real scalable optical computing system, even better designs are needed.
We hope that the open-source implementation of the bandit-like environment serves as a benchmark for the development of new algorithms that can discover these designs.



%% file: sections/06_supplementary.tex
\beginSupplementaryMaterials

\section{Algorithm Details}

\subsection{Structural Priors through Positional Encoding}
A major advantage of expressing the optimization problem as a multi-agent problem is the ability to introduce prior knowledge about the physical structure of the design and thus action space.
We use an observation mapping $O : \{1, \ldots, N\} \rightarrow \mathcal{O}$ that encodes this structure as a positional encoding \citep{vaswani2017attention}.
Specifically, for a discrete design space of size $N = \lvert X \rvert \cdot \lvert Y \rvert \cdot \lvert Z \rvert$ and $b$ bands, $O$ is defined as
\begin{align}
    O(n) &= \begin{bmatrix}
        f(x(n)) \\
        f(y(n)) \\
        f(z(n))
    \end{bmatrix}\, \text{, with} \\
    f(a) &= \begin{bmatrix}
        a \\ \operatorname{sin}(2^0 \cdot \pi \cdot a) \\ \ldots \\ \operatorname{sin}(2^{b-1} \cdot \pi \cdot a) \\ \operatorname{cos}(2^0 \cdot \pi \cdot a) \\ \ldots \\ \operatorname{cos}(2^{b-1} \cdot \pi \cdot a)
    \end{bmatrix}\, \text{,}
\end{align}
where $x : \{1, N\} \rightarrow [-1, 1]$, $y : \{1, N\} \rightarrow [-1, 1]$, and $z : \{1, N\} \rightarrow [-1, 1]$ are functions returning the position of an agent among the 3D grid axis, normalized to range $[-1, 1]$. We use positional encodings as they improve sample efficiency compared to regular multi-layer perceptrons, especially when applied to the critic \citep{yang2022overcoming,schier2023learned}.

\begin{figure}[h]
    \centering
    \includegraphics[width=0.8\linewidth]{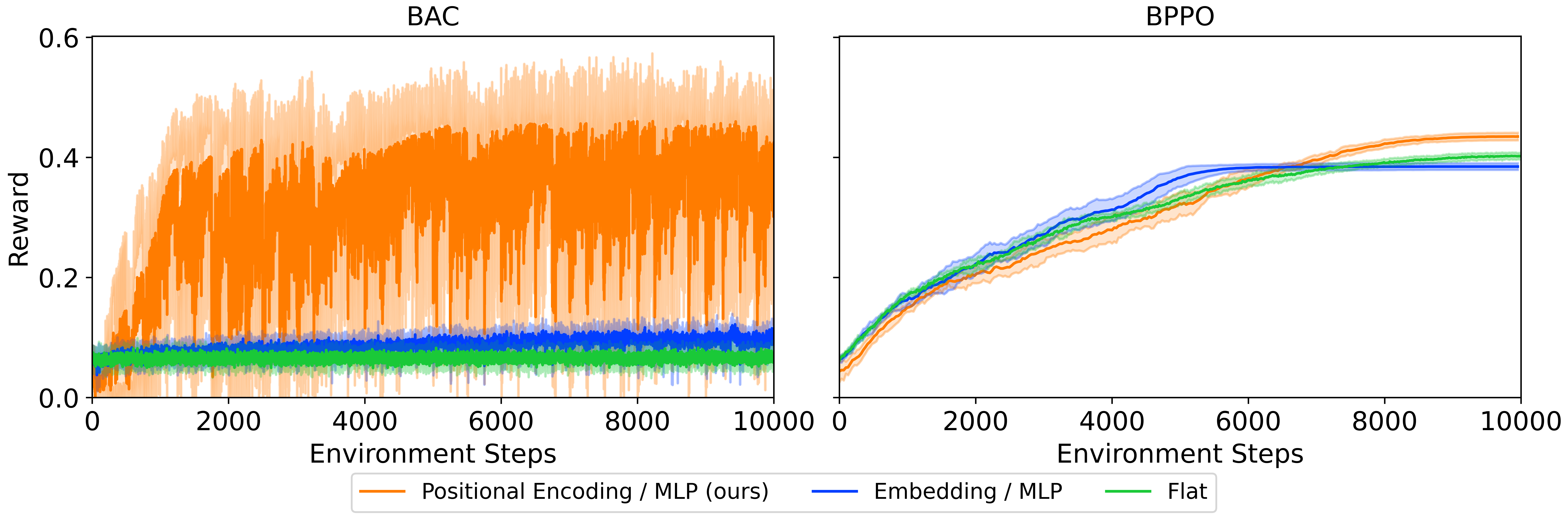}
    \caption{
    Performance of BAC and BPPO with  and without a structural prior through the positional encoding. 
    The experiment is performed on the testing environment presented in section \ref{sec:appendix_gol}.
    Both algorithms achieve a better final design when agents are conditioned on the positional encoding.
    }
    \label{fig:appendix_pe}
\end{figure}

The advantage of using such a positional encoding compared to an $N$-dimensional embedding layer is shown in \cref{fig:appendix_pe}.
We also compare to a flat actor, which is a simple learnable vector with one entry per agent or voxel, describing its activation probability.
This represents a policy under the single agent paradigm acting on the joint action space.
Both BAC and BPPO were trained with agents conditioned on the positional encoding or an equivalently sized embedding layer.
The embedding layer does not contain any information about the structure of the 2D design space, which impedes performance.
Interestingly, BPPO is more robust than BAC to the lack of a positional prior. This may be explained by the common observation that actor-critic algorithms are very sensitive to architecture changes to the critic rather than the policy \citep{bhatt2024crossq} and our BPPO implementation not using a value network.

\subsection{Independent Q-Learning (IQL)}
A simple form of multi-agent learning in discrete action spaces is the independent application of Q-Learning to each agent \citep{tan1993multi, rutherford2023jaxmarl,SchSch2025}.
In our case, payoffs instead of state-action-values are independently learned.
The critic $C_{\psi} : \mathcal{O} \times \{0, 1\} \rightarrow \mathbb{R}$, parameterized by $\psi$, estimates the joint payoff given the action of a single agent. The critic is optimized by gradient descent on
\begin{align*}
    J_C(\psi) = \operatorname*{\mathbb{E}}_{((a_1, \ldots, a_n), r) \sim \mathcal{D}} \left[\frac{1}{n}\sum_{i=1}^n (C_{\psi}(O(i), a_i) - r)^2 \right] \, \text{,}
\end{align*}
where $\mathcal{D}$ is a buffer storing previously collected experiences of joint actions $(a_1, \ldots, a_N)$ and rewards $r$.
The implicit greedy policy for evaluation is $\mu_{\psi}(i) := \operatorname{arg}\operatorname{max}_{a \in \{0,1\}}C_{\psi}(O(i), a)$. 
In order to carry out exploration during training, an $\epsilon$-greedy strategy is employed.
Because the critic estimates global reward from local actions, the non-stationary actions of all other agents are indistinguishable from noise.
Whenever the behavior of other agents in replayed experiences differs from the current implicit policy, either because the implicit policy has changed or because the exploratory policy was used, the global reward signal becomes biased.
Therefore, IQL has no convergence guarantees, but we can reduce the bias of stale experiences by using a small buffer size.

\subsection{Game of Life Hyperparameter Optimization}
\label{sec:appendix_gol}

We implemented a simple testing environment for hyperparameter optimization, which is very quick to evaluate, but still represents a similar structure to the bandit setting of PIC component design.
To this end, we implemented an environment based on Conways game of life \citep{gardner1970mathematical}.
In this game, a 2D grid of cells is simulated, which can be dead or alive.
At each step, a cell that is alive and has either two or three neighbors survives.
Cells with more than three or less than two neighbors die of over- or under-population, respectively.
Additionally, dead cells with exactly three neighbors become alive.
The actions in this game determine the starting configuration for the game of life.
The goal is finding a design, which has as many cells alive as possible, but is stable such that as few cells as possible change in a single step of the game.
Therefore, performance is measured as the difference between the ratio of alive cells and the ratio of changed cells after a single game step.

\begin{figure}[tbh]
    \centering
    \begin{subfigure}{0.48\textwidth}
        \centering
        \begin{minipage}{0.48\textwidth}
            \centering
            \includegraphics[width=\textwidth]{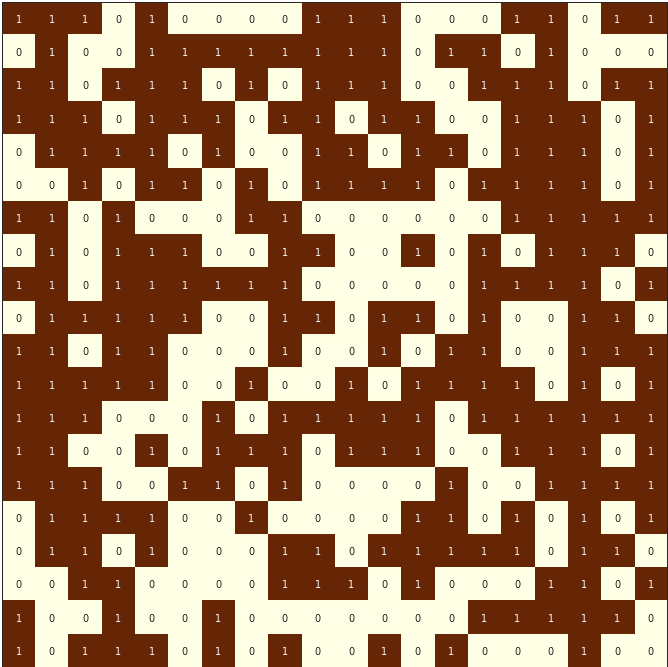}
        \end{minipage}%
        \hfill
        \begin{minipage}{0.48\textwidth}
            \centering
            \includegraphics[width=\textwidth]{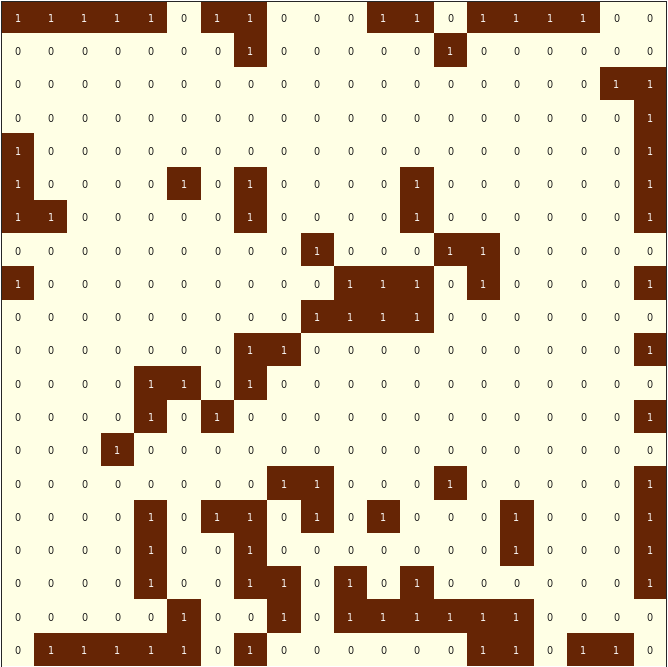}
        \end{minipage}
        \caption{Random Design}
    \end{subfigure}%
    \hfill
    \begin{subfigure}{0.48\textwidth}
        \centering
        \begin{minipage}{0.48\textwidth}
            \centering
            \includegraphics[width=\textwidth]{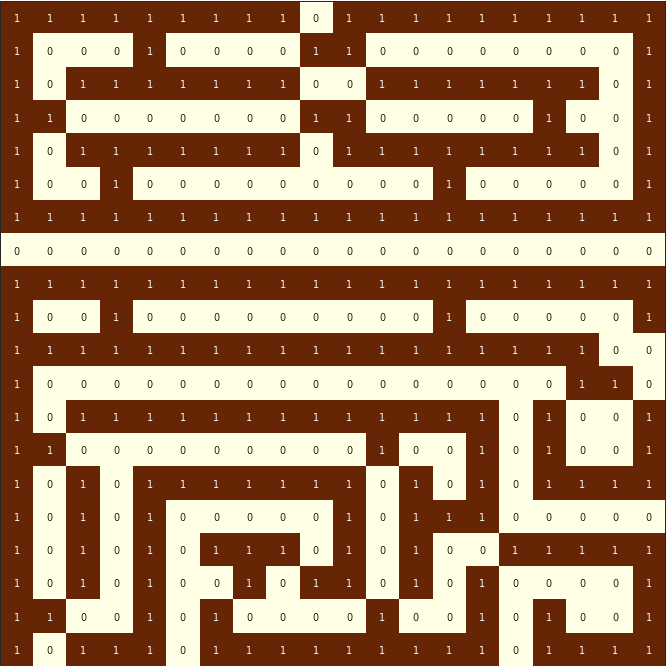}
        \end{minipage}%
        \hfill
        \begin{minipage}{0.48\textwidth}
            \centering
            \includegraphics[width=\textwidth]{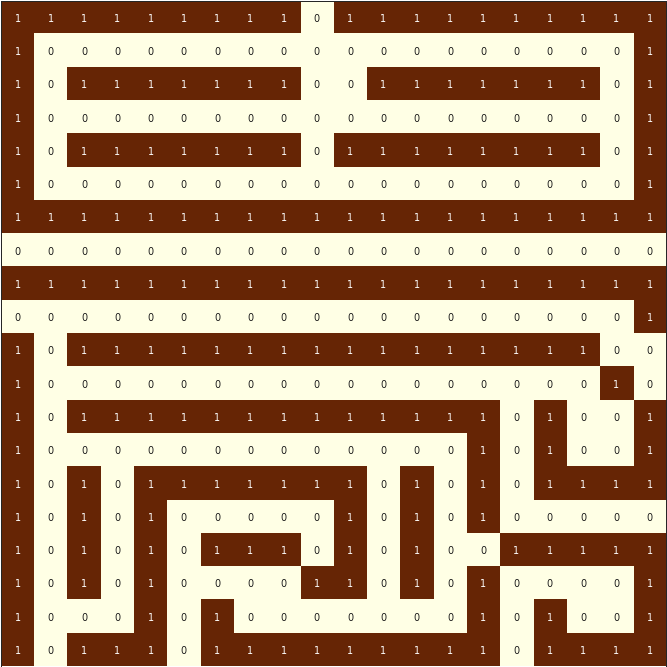}
        \end{minipage}
        \caption{Optimized Design}
    \end{subfigure}
    \caption{Example designs for the game of life environment. In (a), a random design and in (b) a design optimized by BPPO are shown. For both (a) and (b), the left image displays the design and the right side the game of life grid after a single step.}
    \label{fig:game_of_life}
\end{figure}

In \cref{fig:game_of_life}, a random design and a design optimized by BPPO are shown.
The random design achieves a score of $0.04$, because many cells change after the single game step.
In contrast, the optimized design has many alive cells and the grid changes only a little after a game step, resulting in a score of $0.51$.

\subsection{Hyperparameters}
\begin{figure}[tbh]
    \centering
    \includegraphics[width=0.5\textwidth]{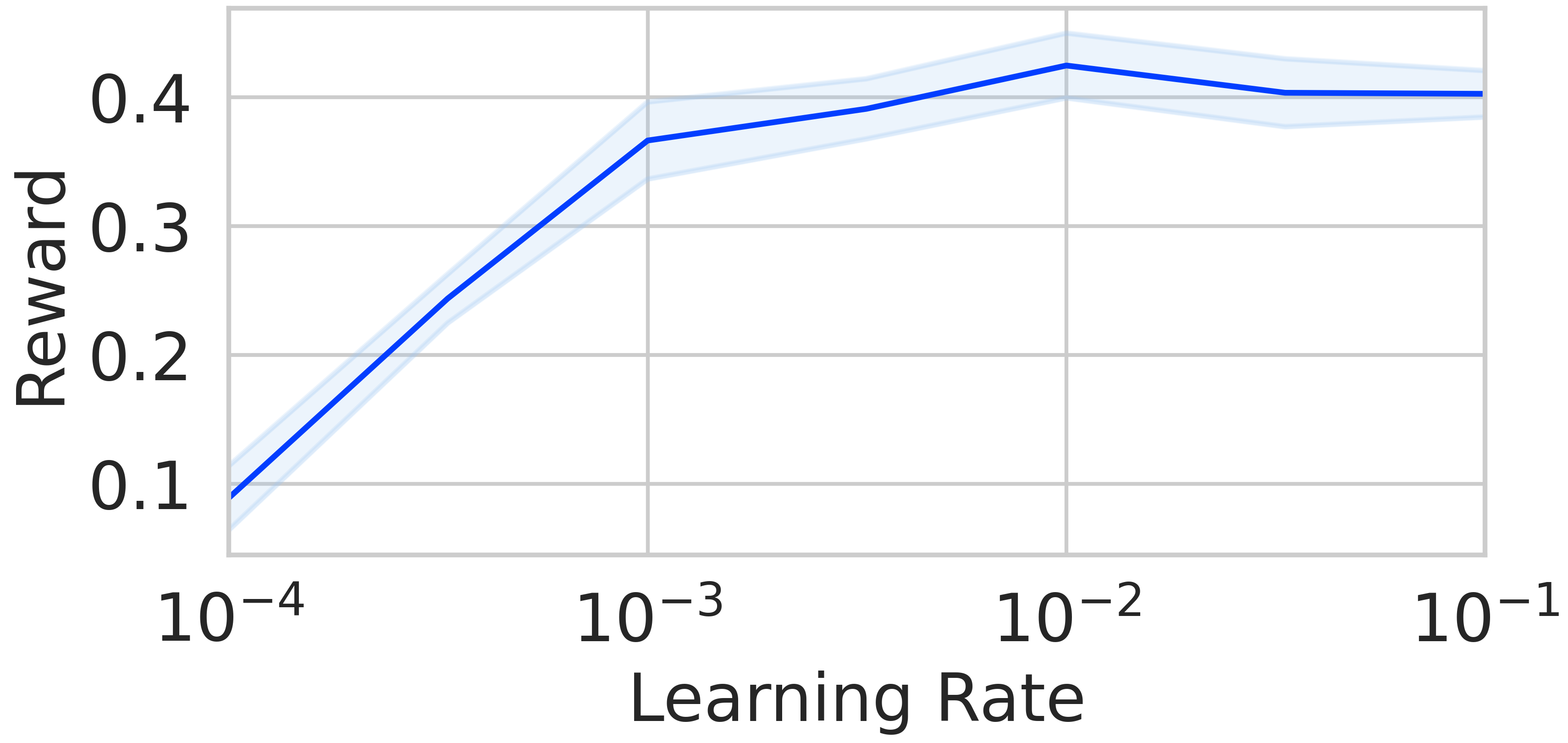}
    \caption{
    Influence of the learning rate hyperparameter on the performance of gradient-based optimization in the silicon coupler environment. Mean and standard deviation are calculated over five seeds.
    }
    \label{fig:grad_lr}
\end{figure}

Using the game of life environment described above, we optimized the various hyperparameters of our algorithms using Optuna \citep{akiba2019optuna}.
Only gradient descent cannot be used in the game of life environment because it is not differentiable.
For gradient-based optimization, we used the adam optimizer \citep{KingmaB14} with nesterov momentum \citep{dozat2016}.
We used the standard parameters of $b_1 = 0.9$, $b_2 = 0.999$ and $\varepsilon = 10^{-8}$.
For the learning rate, we used a cosine scheduling with linear warmup \citep{LoshchilovH17}.
We performed a sweep over the peak learning rate parameter using the silicon coupler environment.
The results of this experiment are shown in \cref{fig:grad_lr}.
We concluded that $0.01$ is the best learning rate, which we used for all the following experiments.

Using DUCT, the problem arises that often all agents select the same actions as DUCT action selection is deterministic and all agents receive the same reward.
Therefore, the first 50 of the 10000 actions during optimization were selected uniformly random.
This ensures that the agents started using the DUCT formula with different initial values.
Moreover, we slightly altered the standard formula by multiplying the exploration term with gaussian noise.
The mean of this normal distribution as well as the exploration constants are hyperparameters, which we optimized in the game of life environment.
The best hyperparameters we found were an exploration factor $c = 0.2145$ and a noise mean of $0.3242$.

\begin{table}[tbhp]
\centering
\begin{tabular}{lcl}
\toprule
\textbf{Parameter} & \textbf{Value} & \textbf{Explanation} \\
\midrule
Solutions per Population & 92 & Number of candidate solutions in each generation \\
\#Parents mating & 8 & Number of solutions selected as parents for breeding \\
Keep parents & 2 & Number of best parents to include in next generation \\
Crossover type & uniform & Genes are randomly swapped with equal probability \\
Mutation type & swap & Mutation by exchanging positions \\
Gene mutation rate & 34\% & Percentage of genes mutated in each offspring \\
\bottomrule
\end{tabular}
\caption{Hyperparameters of the evolutionary algorithm.}
\label{tab:hp_ea}
\end{table}

For evolutionary algorithms, we used the PyGad library \citep{gad2024pygad}.
The hyperparameter optimization resulted in the values listed in \cref{tab:hp_ea}.

\begin{table}[tbhp]
\centering
\begin{tabular}{lccc}
\toprule
\textbf{Parameter} & \textbf{IQL} & \textbf{BAC} & \textbf{BPPO} \\
\midrule
\#Sincos-Bands & 8 & 8 & 8 \\
Learning rate & $10^{-3}$ & $10^{-3}$ ($10^{-4}$) & $10^{-4}$\\
Batch size & 32 & 32 & $10^4$ \\
Buffer size & 200 & $10^4$ & 32 $\times$ \#Agents \\
\#Hidden layer & 2 & 2 & 4 \\
Hidden Dim. & 64 & 128 (256) & 126\\
\bottomrule
\end{tabular}
\caption{Common hyperparameters of the IQL, BAC and BPPO algorithms. For BAC, values in parentheses is for the critic, which is separate from the actor. BAC trains on every simulation result, such that the maximum buffer size is the number of simulations. BPPO performs 32 simulations between gradient updates, such that the number of data collected during rollouts is the number of agents multiplied by 32.}
\label{tab:hp_common}
\end{table}

The IQL, BAC, and BPPO algorithm all used neural networks with a positional encoding as input.
The hyperparameters used by all these algorithms are shown in \cref{tab:hp_common}.
All of the algorithms use an MLP architecture.
For IQL, we use a linear scheduling of the epsilon greedy exploration.
Starting from $\varepsilon = 1$, the random probability is annealed to $\varepsilon = 0.05$ in the first 60\% of the optimization and then is constant.
For BPPO, we used a epsilon clip value of $\varepsilon_{\text{clip}} = 0.4978$ and an entropy loss coefficient of $0.005759$.
BPPO performed 66 gradient updates between rollouts.

For BAC we used the following hyperparameters as shown in Algorithm \ref{alg:bac}.
In our experiments, we used 128 critic gradient steps $U$, 1024 policy gradient steps $G$ and a batch size $B$ of 32. 
Furthermore, at the beginning of optimization, we collect $B$ experiences using uniformly sampled actions to start training the critic on a sufficiently filled experience buffer.
We mask the gradients of 95\% of agents for the optimization of the policy to maintain stochasticity ($m$).
Furthermore, to prevent loss of plasticity in the critic during optimization \citep{nikishin2022primacy}, we reinitialize the critic every 250 steps.
After reinitialization, we perform 512 times as many critic gradient steps as normal to quickly minimize the regression error again.

\section{Full Training Results}

\begin{figure}[tbh]
    \centering

    \subfloat[Si-Coupler]{
        \includegraphics[width=0.46\textwidth]{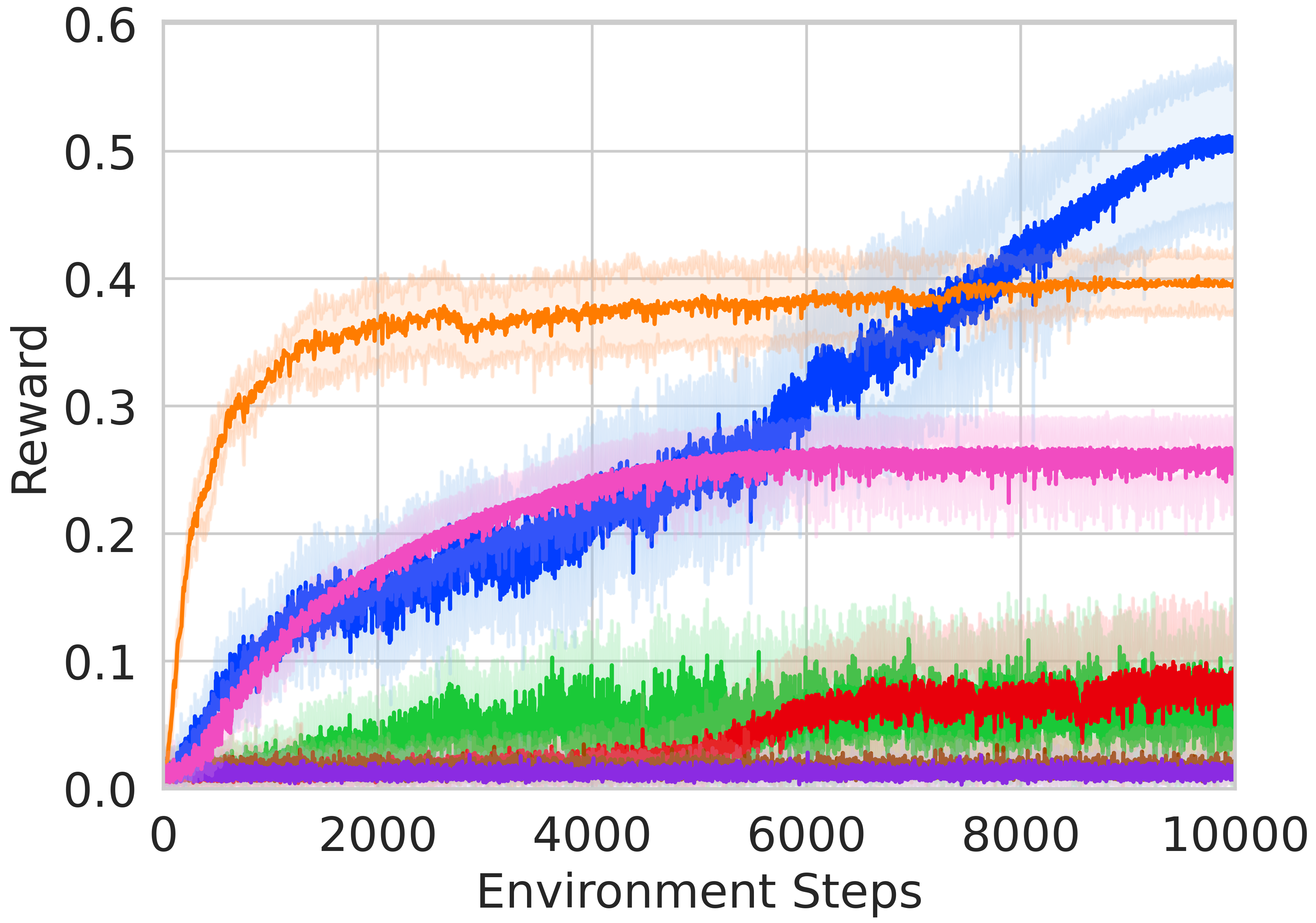}
    }
    \hfill
    \subfloat[P-Coupler]{
        \includegraphics[width=0.46\textwidth]{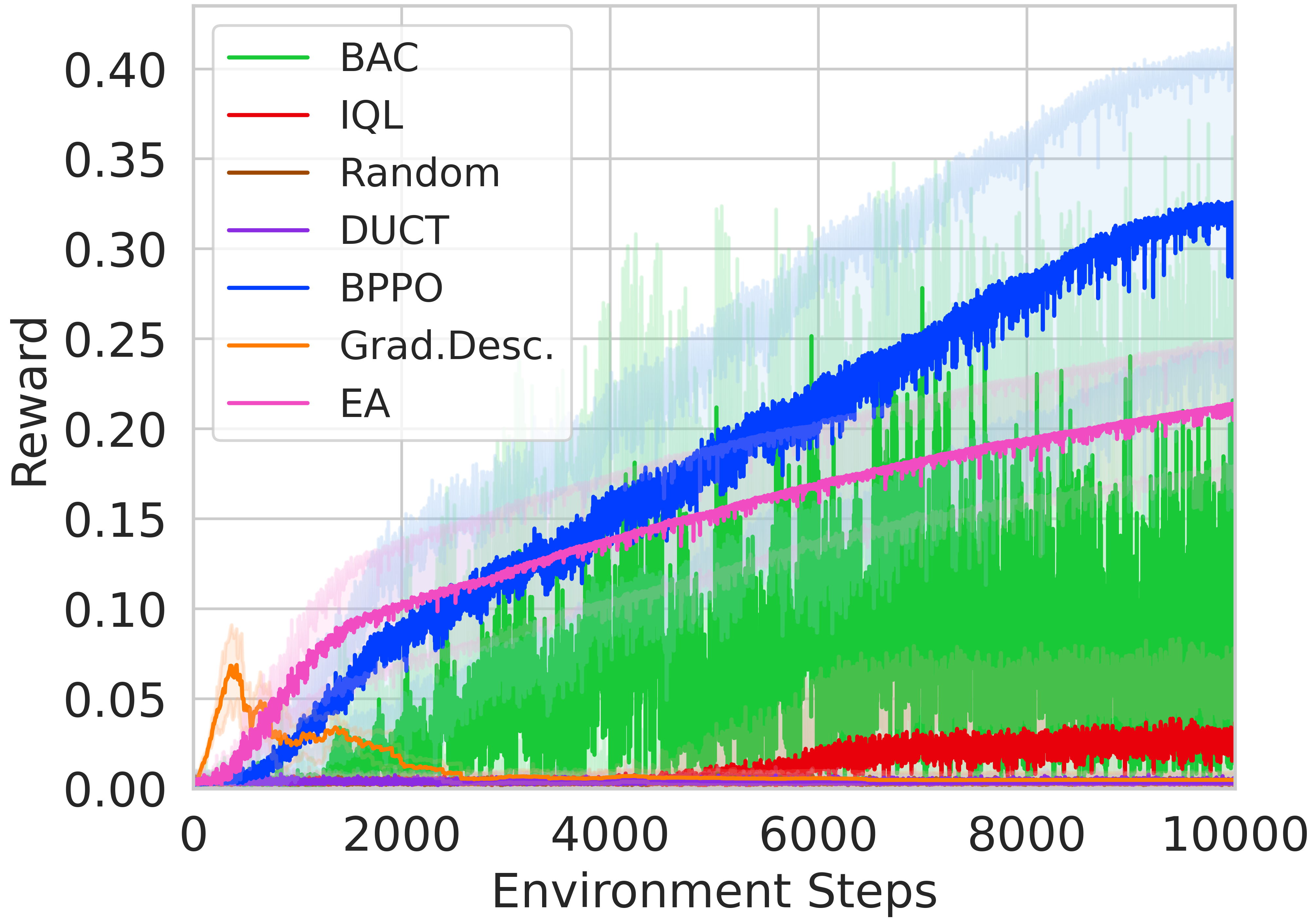}
    }
    
    \subfloat[Si-Corner]{
        \includegraphics[width=0.46\textwidth]{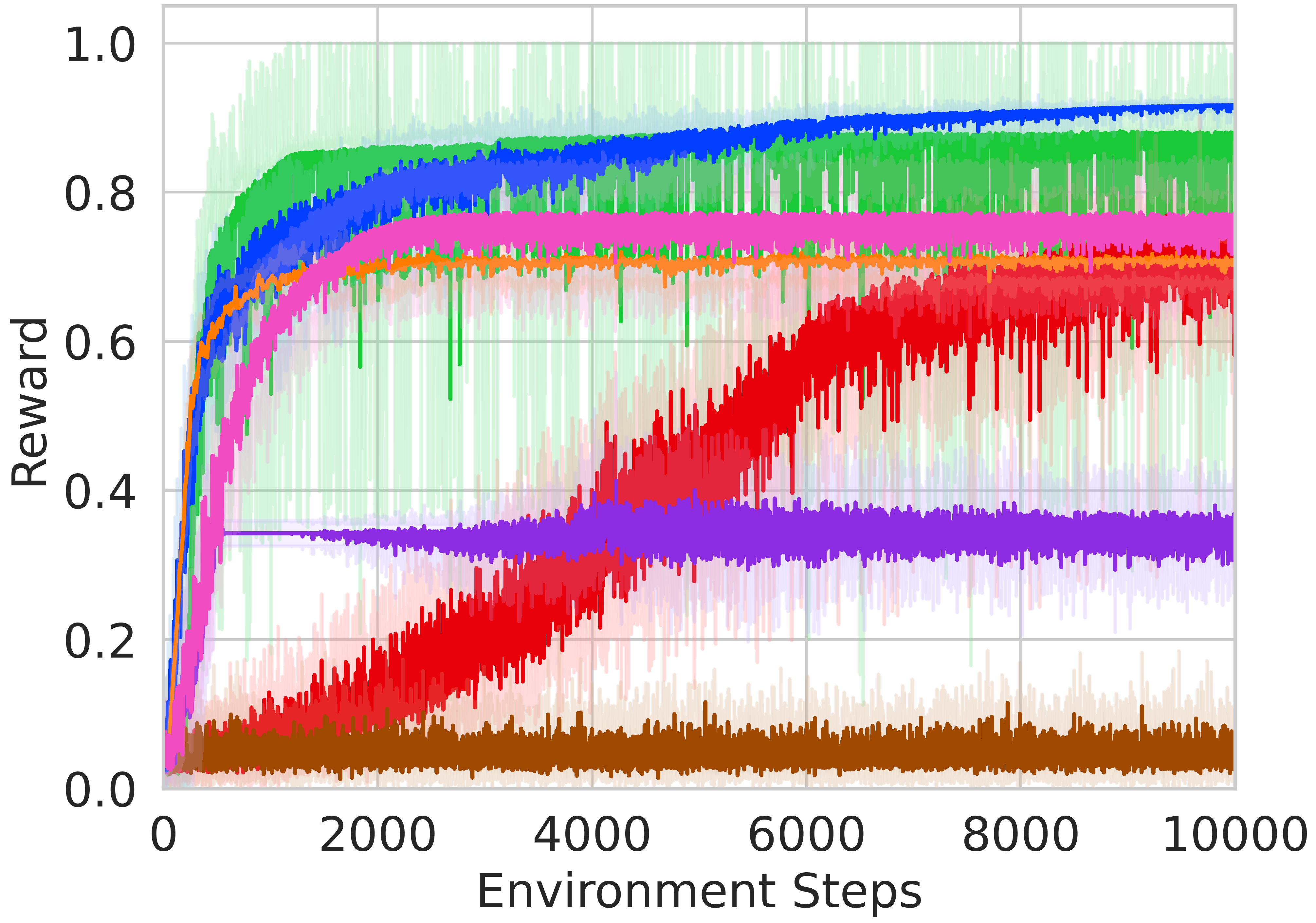}
    }
    \hfill
    \subfloat[P-Corner]{
        \includegraphics[width=0.46\textwidth]{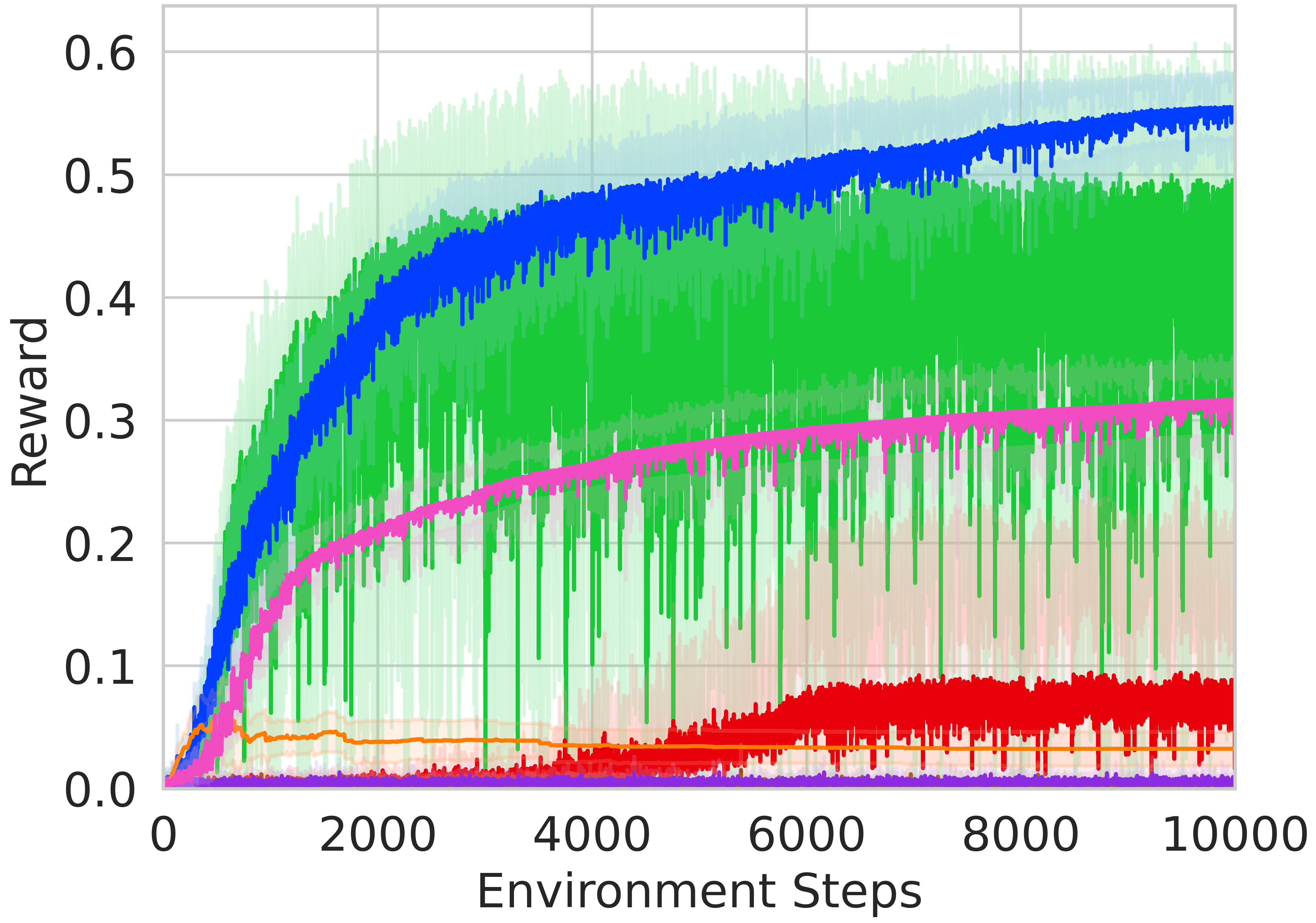}
    }
    
    \subfloat[Si-Vecmul-2]{
        \includegraphics[width=0.46\textwidth]{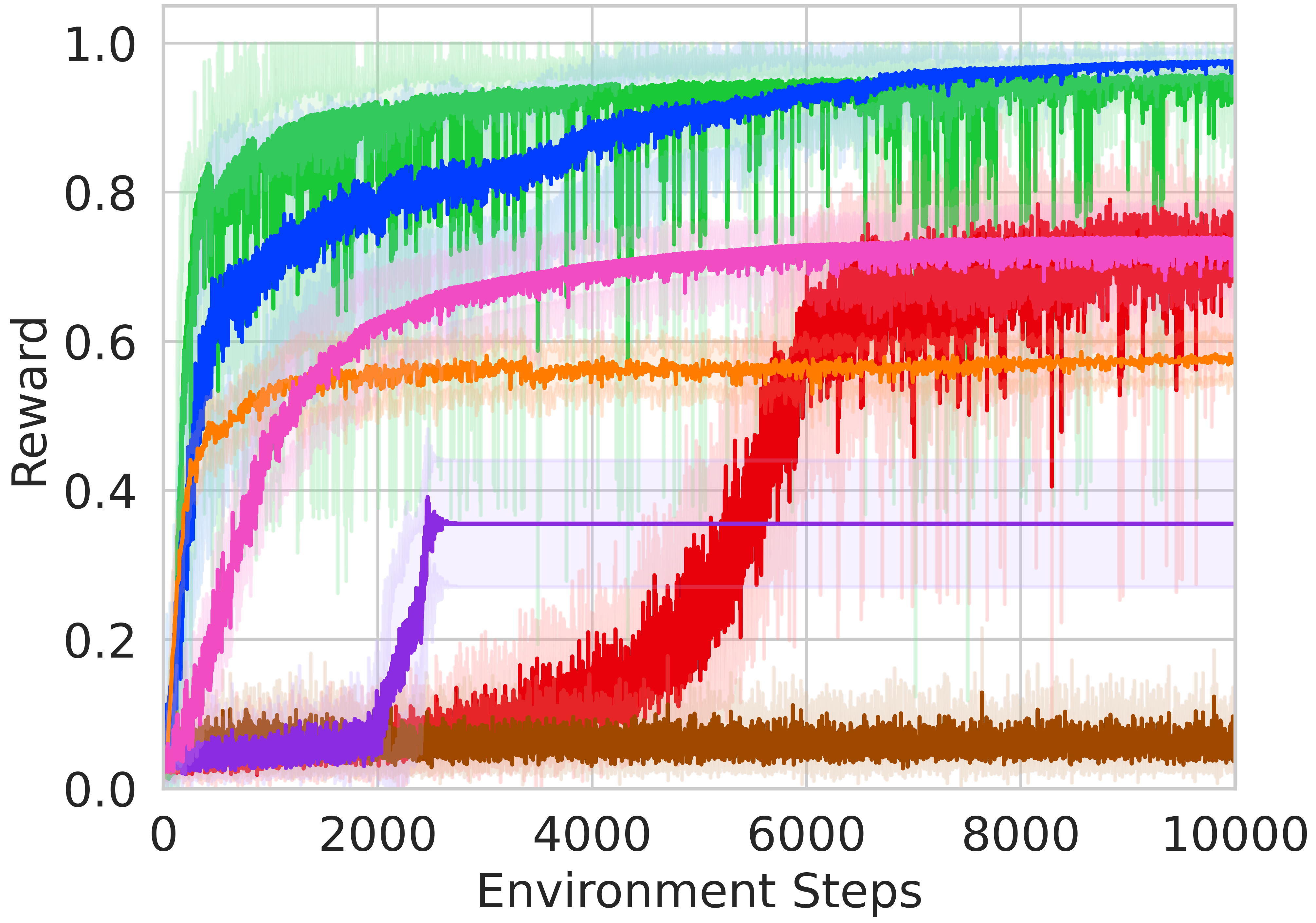}
    }
    \hfill
    \subfloat[P-Vecmul-2]{
        \includegraphics[width=0.46\textwidth]{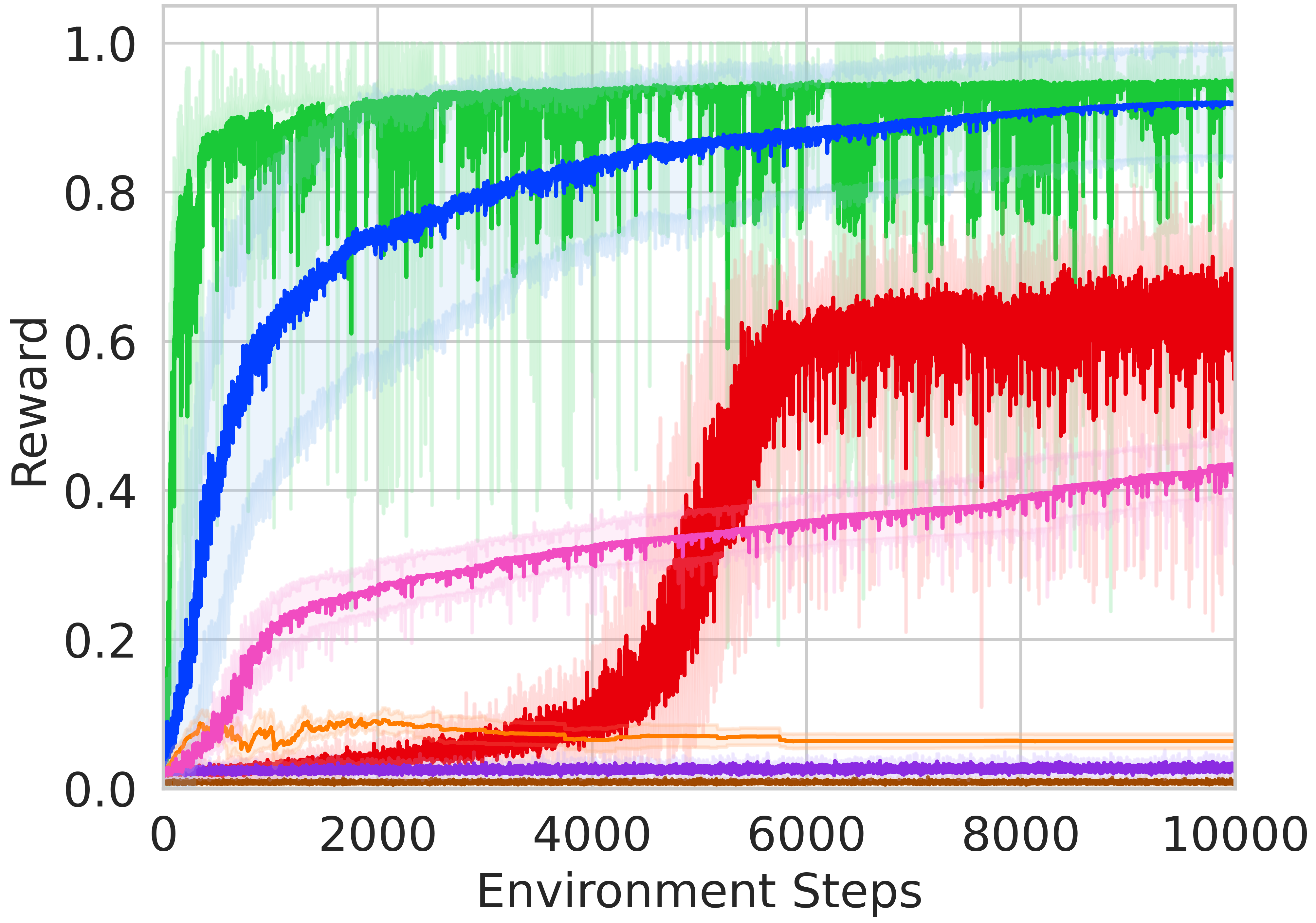}
    }
    
    \subfloat[Si-Vecmul-5]{
        \includegraphics[width=0.46\textwidth]{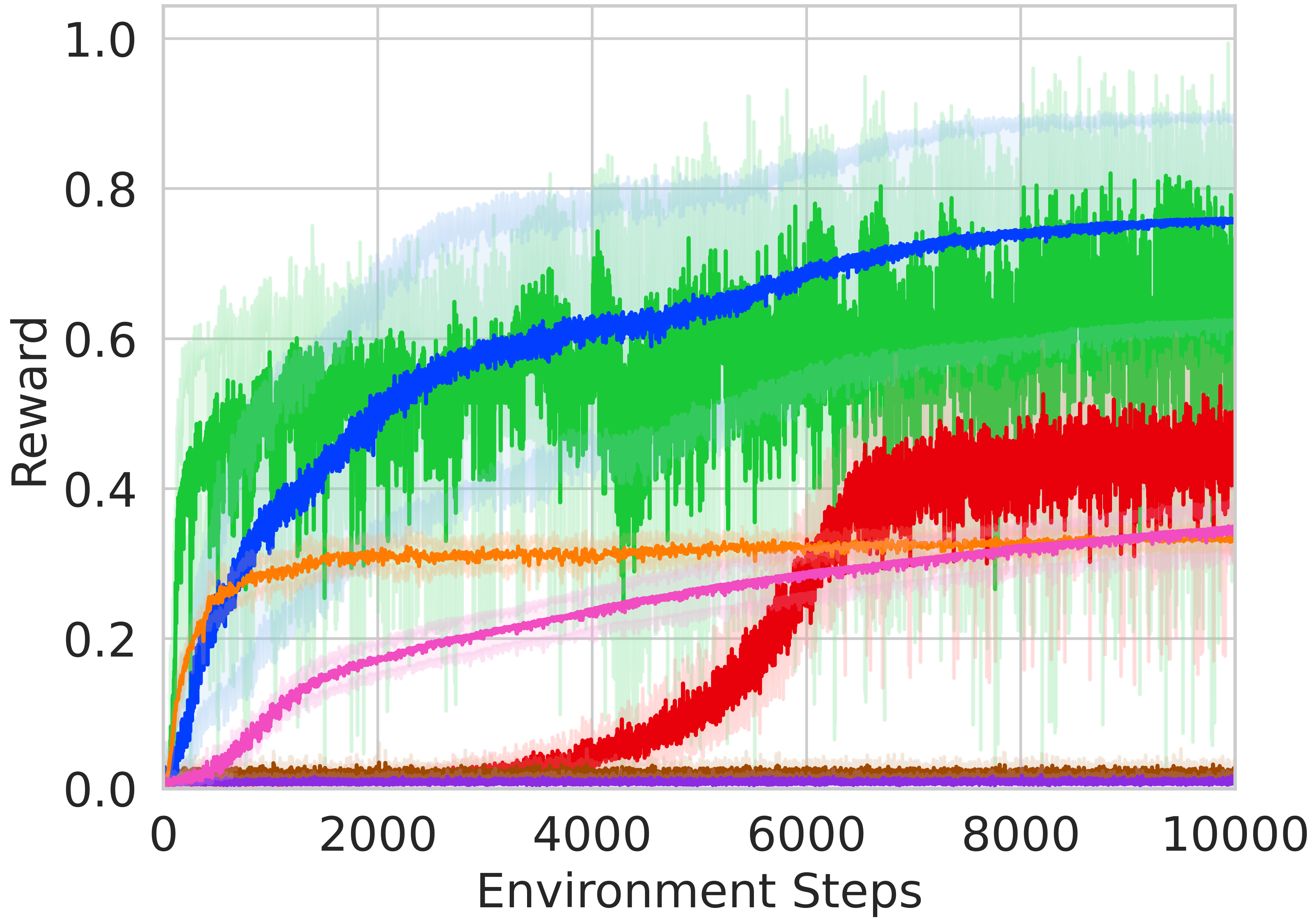}
    }
    \hfill
    \subfloat[P-Vecmul-5]{
        \includegraphics[width=0.46\textwidth]{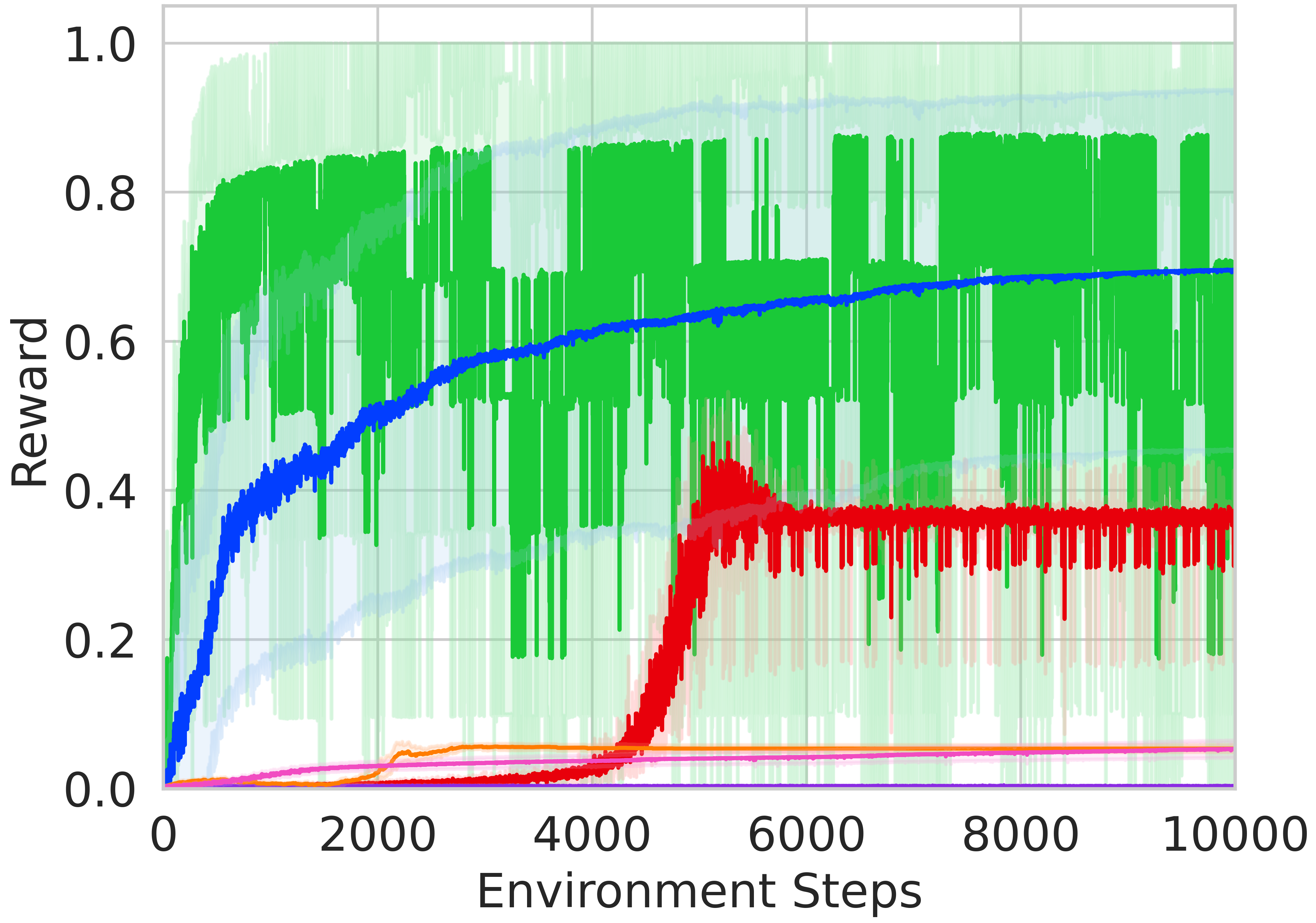}
    }

    \caption{Learning curves for algorithms tested in our work in the different environments. Gradient descent performed fewer steps than the other algorithms because gradient computation required time. In this plot, the x-axis for gradient descent is stretched to allow a comparison with the other algorithms. Mean and standard deviation are calculated over five seeds.}
    \label{fig:trainig_full}
\end{figure}

In \cref{fig:trainig_full}, the full training curves for all environments are shown.
The first interesting observation is that BAC has instable training dynamics, because the critic is often retrained.
In some environments, this may actually be beneficial as it increases exploration.
Since BAC uses all previously collected data to regularly retrain the critic, instable training dynamics can never stop training progress completely.
In contrast, BPPO has smooth training dynamic with a continuously increasing reward during training.
However, in some environments, such as P-Vecmul-5, the variance between different seeds is high.

In the polymer coupler environment, the training collapse of gradient descent can be seen.
In the first few environment steps, gradient descent is able to increase the reward quickly.
However, at some turning point, structures floating in the air or enclosed cavities are removed by the mapping from latent parameters to a physically valid design.
This leads to a large distance between the latent parameters and the actual design used in simulation, which introduces gradient errors.
During optimization, gradient descent is never able to recover from these errors and consequently the performance goes to zero.
In the other environments, the reward achieved by gradient descent also increases quickly in the first few training steps
However, the optimization quickly gets stuck in a local minimum, where the reward remains relatively constant for the rest of the training.

\section{Environment Details}

\begin{figure}[tbh]
    \centering
    \subfloat[P-Corner]{
        \includegraphics[width=0.48\textwidth]{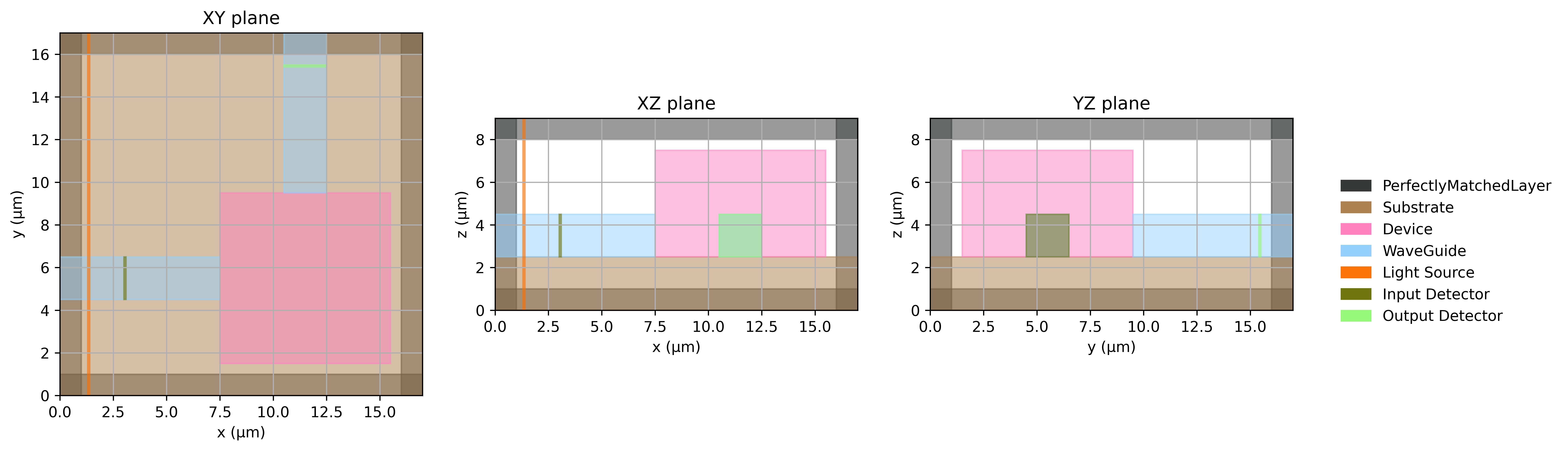}
    }
    \hfill
    \subfloat[P-Coupler]{
        \includegraphics[width=0.48\textwidth]{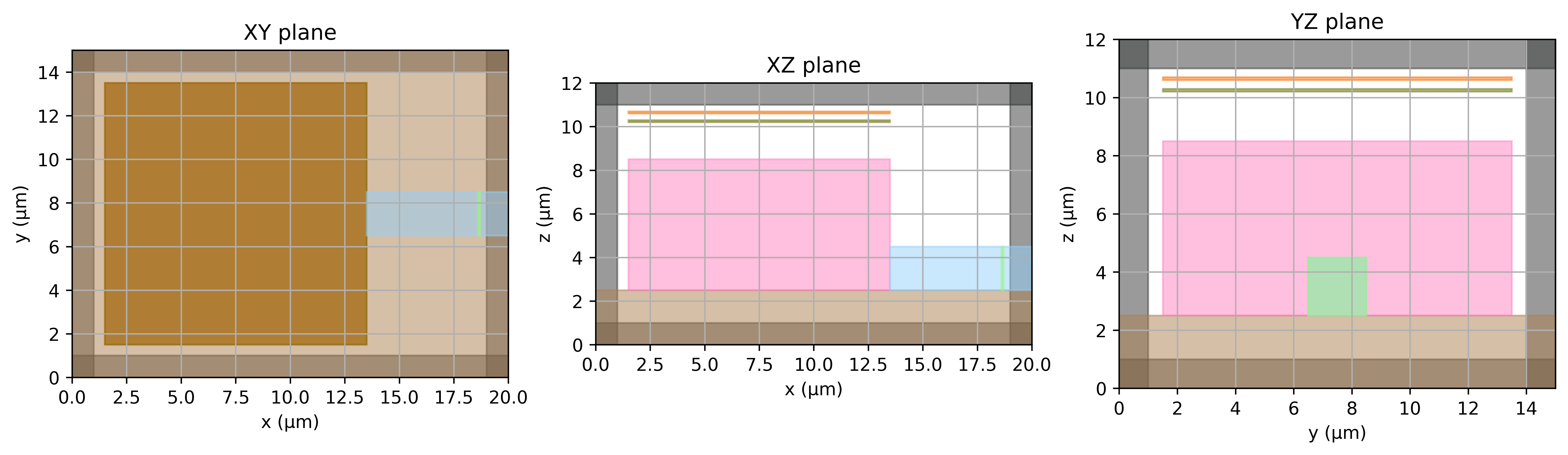}
    }
    
    \subfloat[P-Vecmul-2]{
        \includegraphics[width=0.48\textwidth]{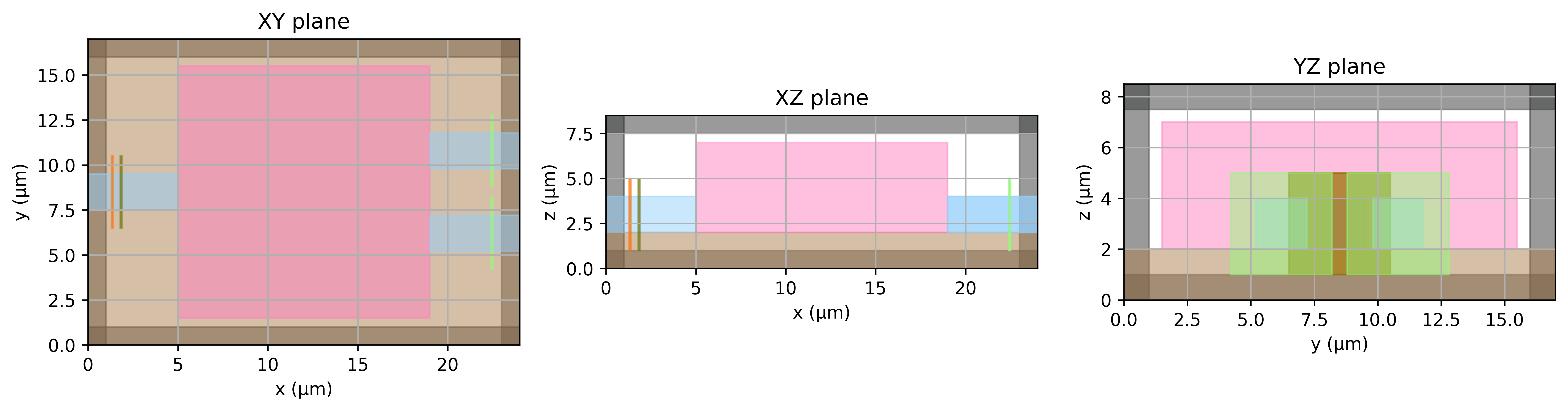}
    }
    \hfill
    \subfloat[P-Vecmul-5]{
        \includegraphics[width=0.48\textwidth]{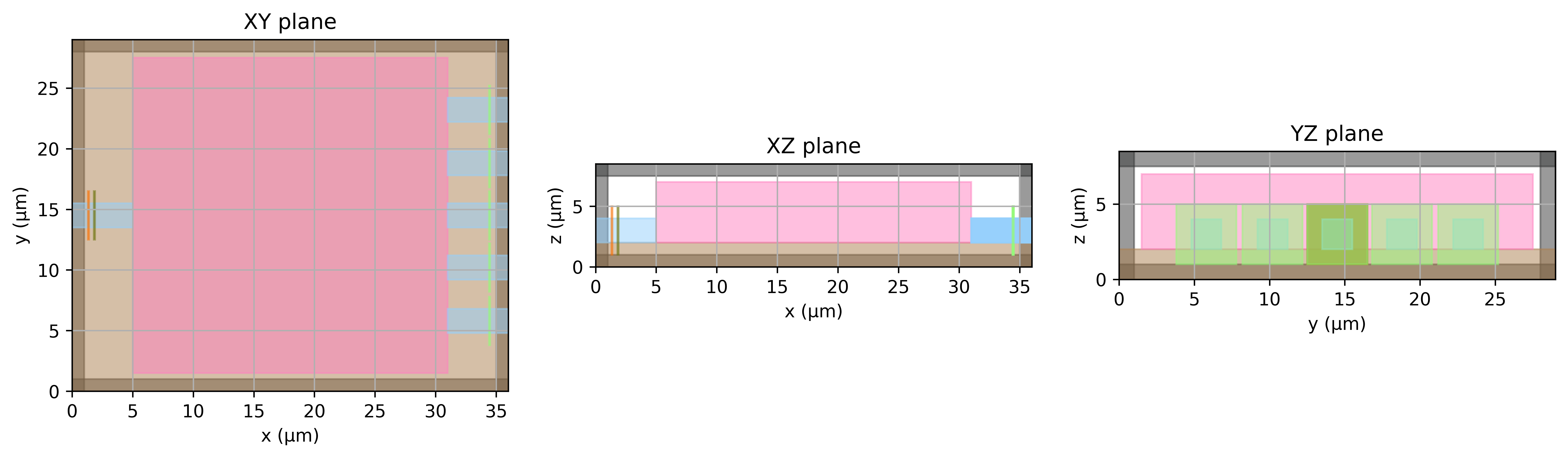}
    }

    \subfloat[Si-Corner]{
        \includegraphics[width=0.48\textwidth]{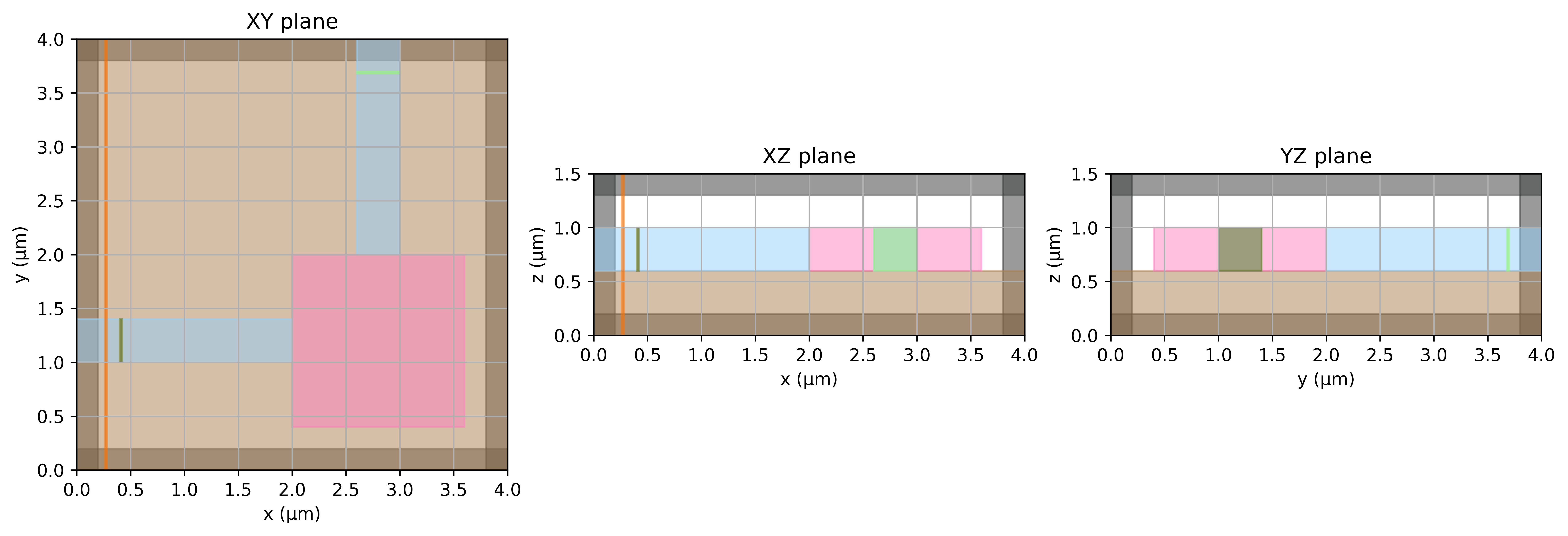}
    }
    \hfill
    \subfloat[Si-Coupler]{
        \includegraphics[width=0.48\textwidth]{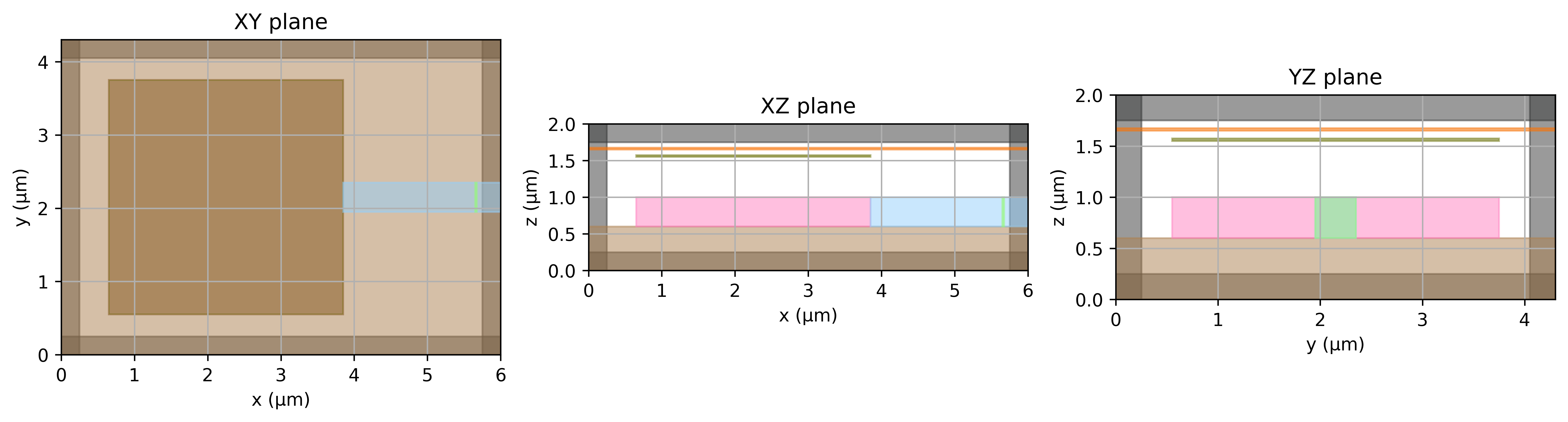}
    }
    
    \subfloat[Si-Vecmul-2]{
        \includegraphics[width=0.48\textwidth]{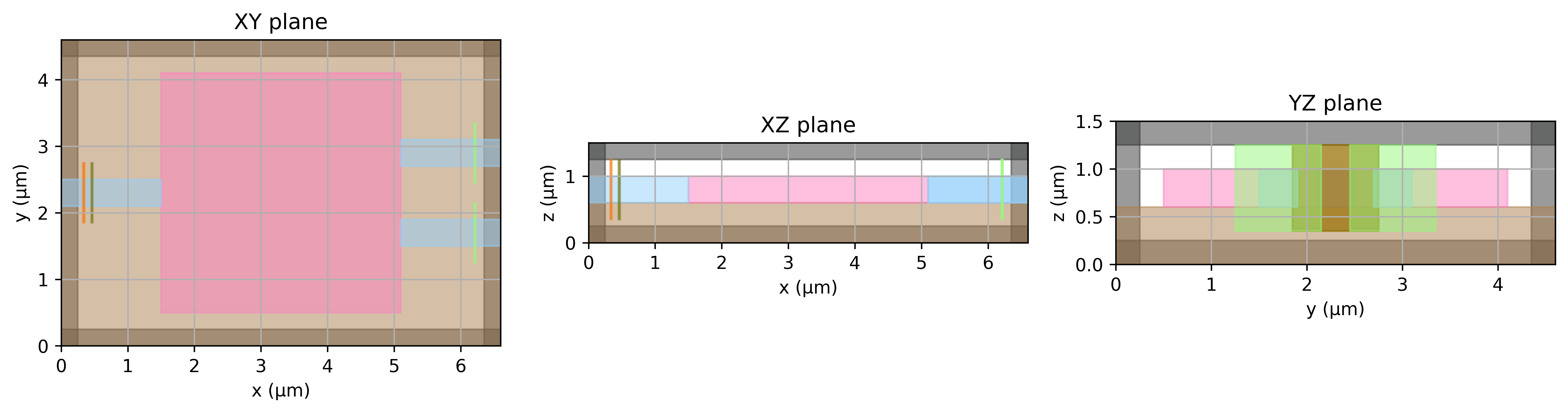}
    }
    \hfill
    \subfloat[Si-Vecmul-5]{
        \includegraphics[width=0.48\textwidth]{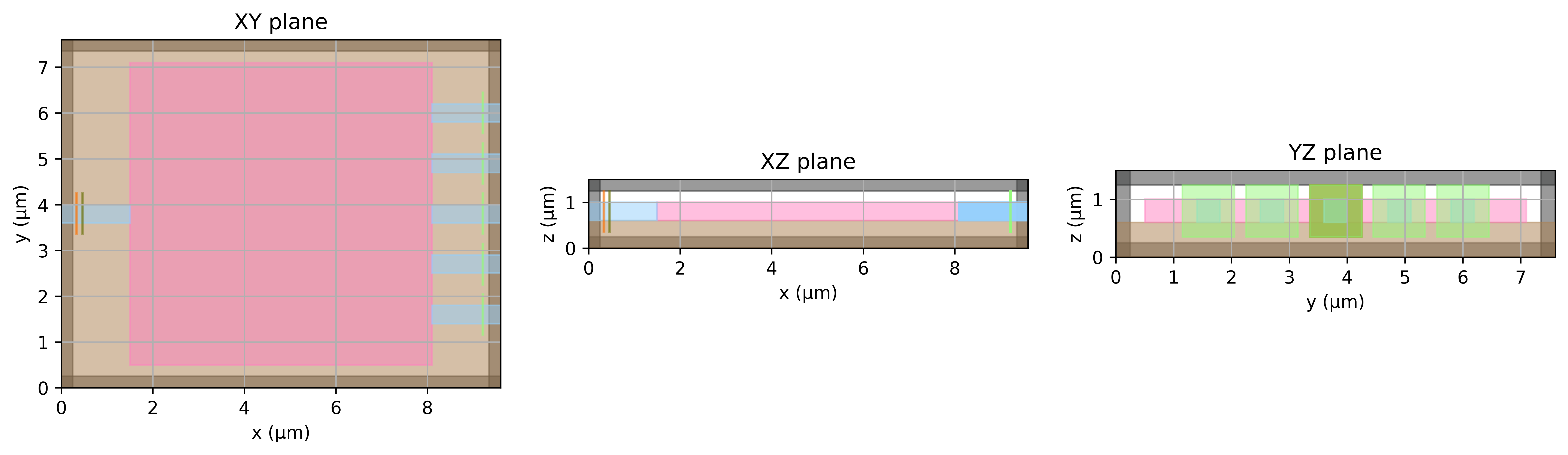}
    }
    
    \caption{Simulation Scenes for the different environments presented in our work.}
    \label{fig:simulation_scenes}
\end{figure}

Our environments perform an electromagnetic simulation to determine the quality of a design.
For simulation, we use the FDTDX software \citep{schubertmahlau2025quantized} written in JAX \citep{jax2018github}, which has been found to be the fastest open-source FDTD software currently available \citep{mahlau2024flexible}.
To induce energy into the simulation with a light source, we use a total-field scattered-field definition \citep{taflove}, which allows unidirectional light input.
We use a light source of wavelength $1550\text{nm}$, which is the standard wavelength for telecommunication.
Reflections at the simulation boundary are prevented through a convolutional perfectly matched layer \citep{cpml}, which absorbs light directed at the boundary.
In \cref{fig:simulation_scenes}, the simulation scenes for all environments are displayed.
For silicon, we assumed a relative permittivity of $12.25$ and for polymer $2.6326$, which corresponds to the material of the ma-N-1400 series \citep{microresisttechnologyProcessingGuidelines}.

\begin{table}[tbhp]
    \centering
    \begin{tabular}{lccccc}
        \toprule
        \textbf{Name} & \textbf{Resolution} & \textbf{Size [µm]} & \textbf{Sim. Time} & \textbf{Sim. Steps} & \textbf{Memory Req.} \\
        \midrule
        Si-Corner & $20$nm & $4 \times 4\ \times 1.5$ & $100$fs & 2623 & $188$MB\\
        Si-Coupler & $25$nm & $6 \times 4.3 \times 2 $ & $125$fs & 2623 & $203$MB\\
        Si-Vecmul-2 & $25$nm & $6.6 \times 4.6 \times 1.5 $ & $106$fs & 2222 & $193$MB\\
        Si-Vecmul-5 & $25$nm & $9.6 \times 7.6 \times 1.5 $ & $156$fs & 3272 & $431$MB\\
        \midrule
        P-Corner & $100$nm & $17 \times 17 \times 9 $ & $200$fs & 1049 & $161$MB\\
        P-Coupler & $100$nm & $20 \times 15 \times 12 $ & $150$fs & 787 & $209$MB\\
        P-Vecmul-2 & $100$nm & $24 \times 17 \times 8.5 $ & $248$fs & 1299 & $210$MB\\
        P-Vecmul-5 & $100$nm & $36 \times 29 \times 8.5 $ & $368$fs & 1929 & $491$MB\\
        \bottomrule
    \end{tabular}
    \caption{Parameters for the electromagnetic FDTD simulations for the different environments. The abbreviation fs is the metrical unit femtosecond. The memory requirements are calculated based on the array sizes of electric and magnetic field, material properties and boundary states, but does not include intermediate values in the FDTD computation.}
    \label{tab:env_params}
\end{table}

In \cref{tab:env_params}, the detailed simulation parameters for the different environments are displayed.
The resolution of the simulation has to be finer than the size of the design voxels to accurately calculate light propagation.
For silicon, the voxel size is $80$nm in the corner environment and $100$nm in the coupler and vecmul-environments.
Therefore, we used resolutions of $20$nm and $25$nm, respectively.
The polymer designs have a voxel size of $500$nm, such that we chose a simulation resolution of $100$nm.
The time discretization, i.e., the time passed per simulation step, is chosen such that the Courant-Friedrichs-Lewy stability conditions \citep{Courant1928berDP} are satisfied.
The time discretization and the simulation time, which is usually in the order of a few hundred femtoseconds, determine the number of simulation steps that need to be performed.
The number of simulation steps is the main factor that influences the computational runtime of the simulation.
Because of the fine resolution, the silicon environments have a higher number of simulation steps than the polymer environments.
With the memory requirements, the maximum number of parallel simulations on a graphics card can be calculated.
However, the true VRAM usage on a graphics card is usually about twice as high as the calculated values in the table because of intermediate computational results in the simulation.
Additionally, a single simulation is already very well parallelized due to its implementation in JAX, such that parallelization of multiple simulations yields little improvement.

The actions in the environment are mapped to a physically valid design for the electromagnetic simulations described above.
The physical validity is determined by the fabrication processes of silicon and polymer, which we describe in more detail here.
Polymer fabrication enables rapid prototyping and small-batch production, while silicon-based PICs provide higher refractive index and smaller feature sizes.

Silicon PICs are fabricated in a subtractive process, where structures are patterned on a waiver.
Common removal techniques include chemical etching \citep{HAN2014271} and focused ion beam milling \citep{HUNG2002256}. 
This fabrication method limits structures to 2D designs with uniform extrusion in the third dimension.
In other words, designs are limited to planar geometries without overhanging structures.
Another important point to consider is the minimum feature size.
Although single-digit nanometer resolution is possible in specialized facilities \citep{cai19}, such precision comes at a significant cost.
For practical implementations, it is more economically viable to utilize fabrication facilities with less accurate machines.
Our experiments use a minimum feature size of $80$nm, which is easily achievable with economically viable fabrication methods.
To ensure strict adherence to this fabrication constraint, we discretize the design space into square voxels of the corresponding size.

In contrast to silicon, polymer can be fabricated into intricate 3D shapes with overhangs and holes.
In the two-photon polymerization (2PP) process \citep{2pp_basics}, liquid monomer resin is placed on a substrate and hardened using a femtosecond laser. 
The process relies on the simultaneous absorption of two near-infrared photons by the photosensitive resin, triggering localized polymerization only at the focal point where photon density is highest.
After polymerization of the desired structures, the remaining liquid monomer is washed away.
Any liquid resin remaining in the design would slowly polymerize over time.
This leads to the fabrication constraint that designs with fully enclosed cavities cannot be produced, which would trap liquid resin that could not be washed away.
In addition, all parts of the design must be connected to the ground, because no structures can float in the air.
This constraint arises from the 3D fabrication capabilities and is not present in 2D silicon designs.
In contrast to silicon, it is not possible to fabricate polymer at nanometer resolution.
In our experiments, we restrict the design space to voxels of $500$nm, which can be achieved by most modern 2PP printers.